\documentclass{article}

\usepackage[final]{neurips_2025}

\usepackage[utf8]{inputenc} 
\usepackage[T1]{fontenc}    
\usepackage{url}            
\usepackage{booktabs}       
\usepackage{amsfonts}       
\usepackage{nicefrac}       
\usepackage{xcolor}         
\usepackage{graphicx}
\usepackage{amsmath}
\usepackage{gensymb}
\usepackage{bbding}
\usepackage[colorlinks]{hyperref}
\usepackage{color}
\usepackage{multirow}
\usepackage[table]{xcolor}
\usepackage{textcomp}
\usepackage{wrapfig}

\usepackage{tikz}
\newcommand{\circlearrow}{\tikz[baseline=-0.5ex, scale=0.3]\draw[->, line width=0.3pt] (0,0) arc (0:-300:0.4);}

\title{
\begin{minipage}{0.1\textwidth}
  \includegraphics[width=\linewidth]{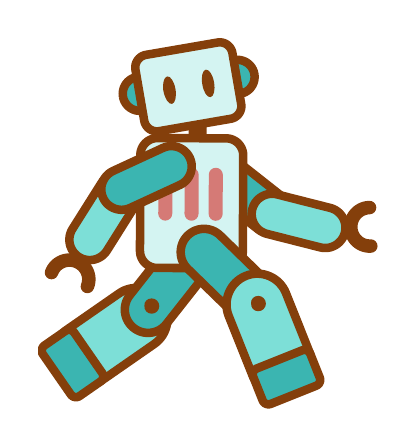}
\end{minipage}
\hfill
\begin{minipage}{0.88\textwidth}
  \textbf{STRIDER: Navigation via Instruction-Aligned Structural Decision Space Optimization}
\end{minipage}
}

\author{Diqi He$^{1}$\footnotemark[1], Xuehao Gao$^{1}$\thanks{Equal contribution.}, Hao Li$^{1,2}$, Junwei Han$^{1,3}$, Dingwen Zhang$^{1}$\thanks{Corresponding author.}\\ 
{$^{1}$}Northwestern Polytechnical University \\
{$^{2}$}Nanyang Technological University  \\
{$^{3}$}Chongqing University of Posts and Telecommunications \\
\href{https://github.com/diqihe666/STRIDER-Nav}{\tt \small{https://github.com/diqihe666/STRIDER-Nav}}
}

\begin{document}

\maketitle

\begin{abstract}
The Zero-shot Vision-and-Language Navigation in Continuous Environments (VLN-CE) task requires agents to navigate previously unseen 3D environments using natural language instructions, without any scene-specific training. A critical challenge in this setting lies in ensuring agents` actions align with both spatial structure and task intent over long-horizon execution. Existing methods often fail to achieve robust navigation due to a lack of structured decision-making and insufficient integration of feedback from previous actions. To address these challenges, we propose STRIDER (Instruction-Aligned Structural Decision Space Optimization), a novel framework that systematically optimizes the agent's decision space by integrating spatial layout priors and dynamic task feedback. Our approach introduces two key innovations: 1) a Structured Waypoint Generator that constrains the action space through spatial structure, and 2) a Task-Alignment Regulator that adjusts behavior based on task progress, ensuring semantic alignment throughout navigation. Extensive experiments on the R2R-CE and RxR-CE benchmarks demonstrate that STRIDER significantly outperforms strong SOTA across key metrics; in particular, it improves Success Rate (SR) from 29\% to 35\%, a relative gain of 20.7\%. Such results highlight the importance of spatially constrained decision-making and feedback-guided execution in improving navigation fidelity for zero-shot VLN-CE.

\end{abstract}

\section{Introduction}

VLN-CE challenges agents to follow natural language instructions to navigate previously unseen 3D environments without any scene-specific training or fine-tuning~\cite{qiao2024open, krantz2020beyond, chen2024constraint, zheng2024towards, liang2023cornav}. This task is a critical benchmark for embodied AI, requiring agents to generalize perception, reasoning, and action across diverse and dynamic scenes~\cite{anderson2018vision, krantz2021waypoint, ma2019regretful, zhu2020vision, song2023llm, chen2021history}. In comparison to discrete VLN tasks~\cite{anderson2018vision, thomason2020vision, ku2020room}, VLN-CE more closely reflects real-world deployment conditions, where agents must process RGB-D inputs and make continuous movement decisions~\cite{hong2022bridging, zhou2024navgpt, jeong2024zero}. As a result, zero-shot VLN-CE pushes the boundaries of language-grounded generalization in embodied navigation~\cite{long2024discuss, zhou2024navgpt, liang2023cornav, chen2024constraint}.

\begin{figure*}[t]
  \centering
  \includegraphics[width=1\linewidth]{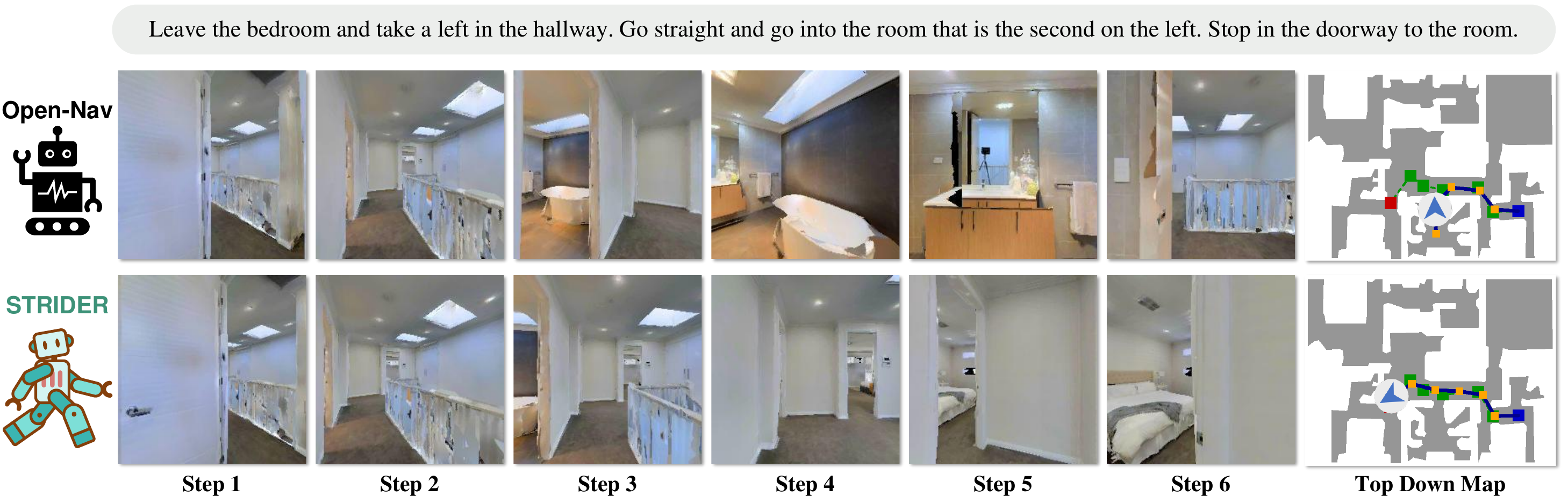}
  \caption{\textbf{Navigation behavior comparison between STRIDER and Open-Nav~\cite{qiao2024open}.} Given the same instruction, Open-Nav demonstrates execution drift, such as prematurely turning away from a hallway, and accumulates deviations over time. In contrast, STRIDER generates trajectories that more accurately follow the intended path and reach the goal region.}
  \label{fig:teaser}
\end{figure*}

A central challenge in VLN-CE lies not only in grounding instructions into perception, but also in ensuring that the agent's behavior remains aligned with the semantic intent of the instruction throughout the navigation process. In unfamiliar environments, agents may correctly understand the instruction yet still exhibit execution drift~\cite{song2023llm, li2023kerm, qiao2024open, chen2021history, hong2022bridging, liang2023cornav}, such as stopping near the target room without entering or prematurely turning away from a hallway, as shown in Fig.~\ref{fig:teaser}. These failures highlight a significant gap between \textit{what the agent understands} and \textit{how it acts}. Our key insight is that effective zero-shot VLN-CE agents must go beyond strong perception and reasoning—--they must operate within an \textit{Instruction-Aligned Structural Decision Space}, a decision space that is explicitly structured by the environment and continuously regulated based on task progress.

However, existing approaches typically rely on learned waypoint predictors or sequence-to-sequence policies that map instructions and visual inputs directly to actions~\cite{qiao2024open, krantz2021waypoint, hong2022bridging, chen2021history}. While effective at modeling local navigability, these models tend to ignore structured representations that capture the global layout or semantic task progression~\cite{guhur2021airbert, hao2020towards, majumdar2020improving, ku2020room, li2023kerm}. Moreover, these methods often operate in an open-loop fashion, inferring each action independently without feedback on prior decisions~\cite{an2024etpnav, chen2022think, wang2023gridmm, long2024discuss, zhou2024navgpt}. This limitation hinders their ability to assess whether actions have brought the agent closer to the goal, leading to deviations from the instruction's intent, especially in complex or ambiguous scenes. While instruction grounding has seen significant progress, insufficient attention has been paid to optimizing the agent's decision space in a manner that aligns both with spatial structure and task instructions.

To address these challenges, we introduce \textbf{STRIDER}, a zero-shot VLN-CE framework built on the principle of \textit{Instruction-Aligned Structural Decision Space Optimization}. Our approach is grounded in the observation that semantic misalignment often arises not from perceptual misunderstanding, but from the inability to maintain alignment with both the spatial structure of the environment and the task's semantic progress over long-horizon execution~\cite{anderson2018vision,ke2019tactical}. STRIDER adopts a modeling-first approach: instead of directly predicting actions from visual inputs and instructions, it focuses on optimizing the agent's decision space by integrating spatial structure and task progress awareness~\cite{tan2019learning}. By embedding spatial layout priors and continuous goal feedback into the agent's decision-making process, STRIDER enables the agent to navigate within paths that are both spatially coherent and semantically aligned with the task. This dynamic adjustment of behavior reduces execution drift and improves instruction fidelity, enhancing both spatial generalization and instruction-level adherence.

STRIDER achieves this optimization through two tightly integrated modules. The \textbf{Structured Waypoint Generator} creates a layout-constrained action space by extracting skeletons from depth-based navigable regions~\cite{chen2022fast, lien2006simultaneous, sharifipour2020skeletonization}. This module ensures that the agent's movement decisions are limited to paths that are both spatially coherent and meaningful, grounded in the environment's structure. The \textbf{Task-Alignment Regulator} continuously monitors task progress and adjusts the agent's behavior accordingly, ensuring actions remain aligned with the instruction-defined goal. It recalibrates behavior whenever deviations are detected, while staying within the spatial constraints defined by the structured action space. Together, these modules optimize the agent's decision space by structuring it with spatial constraints and regulating it according to task progress, ultimately enabling more efficient and semantically faithful navigation.

Our contributions are summarized as follows:
\begin{itemize}
    \item We propose \textbf{STRIDER}, a zero-shot VLN-CE framework based on the principle of \textit{Instruction-Aligned Decision Space Optimization}. STRIDER optimizes the agent's decision space by integrating spatial structure and task feedback, enabling more coherent and instruction-faithful navigation behavior.
    
    \item STRIDER consists of two tightly coupled modules that jointly optimize the agent's decision space:
    (1) A \textbf{Structured Waypoint Generator} that constructs a layout-constrained planning space from depth-based skeletons, embedding spatial priors into the action space;
    (2) A \textbf{Task-Alignment Regulator} that monitors semantic progress across steps and adjusts behavior accordingly, ensuring actions remain aligned with instruction goals.

    \item We evaluate STRIDER on two standard zero-shot VLN-CE benchmarks, R2R-CE and RxR-CE, where it consistently outperforms strong baselines on core navigation metrics, demonstrating the benefit of structuring and regulating the agent's decision space.
\end{itemize}

\section{Related Work}

\subsection{Vision-and-Language Navigation}

Significant progress has been made in Vision-and-Language Navigation (VLN), especially in continuous settings~\cite{qiao2024open, krantz2021waypoint, krantz2020beyond, hong2022bridging, ma2019self, anderson2018vision}. VLN approaches can be broadly categorized into supervised learning with environment-specific training and zero-shot generalization. Supervised methods, such as imitation learning and reinforcement learning~\cite{wang2019reinforced}, rely on human-annotated trajectories to train navigation policies~\cite{an2024etpnav, ma2019self, fried2018speaker, zhang2024navid, zhang2025mapnav, hong2020recurrent}, often enhancing performance through visual-language alignment, attention modules, auxiliary tasks~\cite{ma2019self, tan2019learning, zhu2020vision, an2022bevbert}, and explicit or implicit 3D scene understanding and reconstruction~\cite{wang2024lookahead, li2024langsurf, li2025dgtr, li2024ggrt, gao2024cosurfgs, li2024gp}. On the other hand, zero-shot methods aim to generalize to unseen scenes without task-specific fine-tuning. These approaches typically leverage instruction tuning, pretrained Vision-Language Models (VLMs) or Large Language Models (LLMs)~\cite{qiao2024open, long2024instructnav, shi2025smartway, wang2024panoptic, sun2024review}. While zero-shot methods show promise, they primarily focus on local observations and immediate actions, lacking global spatial awareness and long-term planning capabilities. Few methods integrate feedback mechanisms to monitor task progress or correct deviations, limiting their adaptability in complex scenarios.

\subsection{Decision Space Optimization}

Decision Space Optimization originates from robotics and autonomous systems, where it involves structuring an agent's decision-making within a "decision space" that integrates spatial constraints, task goals, and feedback from the environment~\cite{lavalle2006planning, toussaint2015logic, kaelbling2013integrated, li2023modular}. This structured decision-making allows agents to balance short-term actions with long-term objectives, improving task execution. In robotics, decision space optimization ensures that actions are spatially coherent and consistent with global goals, aiding navigation and manipulation in dynamic environments~\cite{tamar2016value, ratliff2009learning, chen2024metric}. 
In the field of VLN, early methods primarily map instructions to actions based on local observations but struggle with long-term planning and generalization to unseen environments~\cite{anderson2018vision, fried2018speaker, ma2019self}. More recent methods have incorporated memory, attention mechanisms, and reinforcement learning to improve decision-making~\cite{wang2019reinforced, hong2020sub, majumdar2020improving}. These approaches also use visual-language alignment for action prediction and introduce curriculum learning for enhanced performance~\cite{zhu2020vision, an2024etpnav, tan2019learning}. However, challenges remain in maintaining global spatial awareness and ensuring long-term task alignment. STRIDER addresses these issues through Instruction-Aligned Structural Decision Space Optimization. It optimizes decision-making by ensuring actions are consistent with both spatial constraints and task goals, enabling better navigation performance over long horizons.

\section{Method}

\begin{figure*}[t]
  \centering
  \includegraphics[width=1\linewidth]{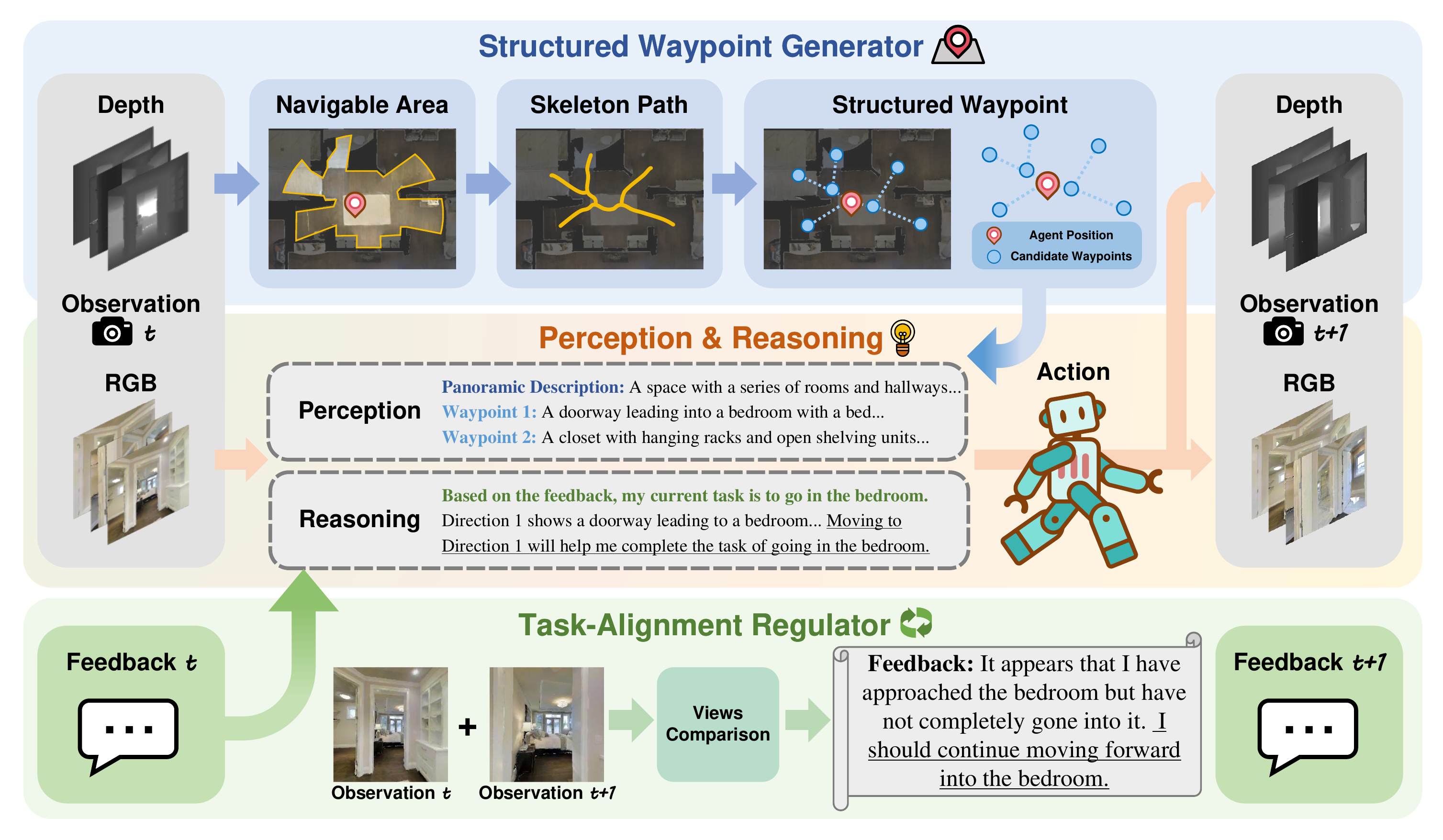}
  \caption{\textbf{Overview of the STRIDER pipeline.} The Structured Waypoint Generator constructs a layout-constrained waypoint space by extracting skeleton paths from navigable depth observations. The agent performs perception and reasoning over visual descriptions and feedback to identify suitable actions in context. To maintain semantic alignment over time, the Task-Alignment Regulator compares current and previous observations and generates feedback that guides the next action.}
  \label{fig:pipeline}
\end{figure*}

In this section, we explore how VLN-CE can be enhanced by structuring the decision space through spatial constraints and integrating task feedback.
We begin with an overview of the VLN-CE task and introduce our proposed STRIDER framework (Sec.~\ref{sec:3.1}).
We then present two core components of our approach:
(1) the use of local perception to construct spatially organized representations that capture region connectivity (Sec.\ref{sec:3.2}), and
(2) the integration of task feedback to monitor subgoal progress and regulate action selection during execution (Sec.~\ref{sec:3.3}).

\subsection{Overview}
\label{sec:3.1}

\textbf{Task Definition.} 
In the task of zero-shot VLN-CE, the agent begins at a specified starting location and must reach a target destination by interpreting both verbal guidance and visual observations. At each timestep, it receives a panoramic 360° view composed of 12 RGB-D images captured at fixed intervals (0°, 30°, ..., 330°). From these observations, the agent predicts a set of waypoints, candidate navigable locations, and selects one to move toward. The episode continues until the instruction is fulfilled or the agent reaches the goal.

\textbf{STRIDER.} 
Our proposed STRIDER framework follows a zero-shot VLN-CE pipeline, as illustrated in Fig.~\ref{fig:pipeline}. At each timestep $t$, the agent receives an RGB-D observation $\textbf{O}_t = (\textbf{I}_t, \textbf{D}_t)$ and a language instruction $L$. The Structured Waypoint Generator processes the depth $\textbf{D}_t$ to extract navigable regions and generates a set of structured waypoints $\mathcal{W}_t$ organized by spatial connectivity.
Subsequently, a pretrained Vision-Language Model (VLM) describes the RGB input $\textbf{I}_t$, attending to visual content in the directions of candidate waypoints $\mathcal{W}_t$ to form a decision space $\mathcal{A}_{t}$.
A Large Language Model (LLM) then reasons over the instruction ${L}$, the current decision space $\mathcal{A}_{t}$, and the task feedback $f_{t}$ generated from the previous step to select a waypoint $w_t^* \in \mathcal{W}_t$ for movement.

After executing the action toward $w_t^*$, the agent receives the next observation $\textbf{O}_{t+1}$. The Task-Alignment Regulator compares $\textbf{O}_t$ and $\textbf{O}_{t+1}$ using the VLM to detect progress toward the instruction goal and generates updated feedback $f_{t+1}$. This feedback is leveraged in the next decision step, completing a closed-loop control cycle grounded in structured perception and task alignment.

\subsection{Structured Waypoint Generator}
\label{sec:3.2}

To optimize the agent's decision space in continuous environments, we generate a layout-constrained set of candidate actions that explicitly reflect the spatial structure of the scene. Rather than relying on local navigability or unconstrained policy outputs, we propose a \textbf{Structured Waypoint Generator} that transforms depth input into a compact topological graph of navigable options. This process consists of three stages: (1) navigable region extraction, (2) topological skeleton abstraction, and (3) structured waypoint selection.

\textbf{Navigable Region Extraction.}
At time step $t$, the agent receives a panoramic RGB-D observation $\textbf{O}_t = (\textbf{I}_t, \textbf{D}_t)$, where $\textbf{D}_t$ denotes multi-view depth input. Then we reconstruct a local point cloud $\mathcal{P}_t \subset \mathbb{R}^3$, where each point $\mathbf{p}_i = (x_i, y_i, z_i)$ represents a 3D coordinate in the agent-centric reference frame. To isolate feasible movement areas, we filter for ground-level points and project them into a 2D top-down plane:
\begin{equation}
\Omega_t = \Pi\left( \left\{ \mathbf{p}_i \in \mathcal{P}_t \;\middle|\; \mathbf{p}_i(z) <\delta_h,\ \|\mathbf{p}_i(x,y)\| < r \right\} \right),
\label{eq:navigable_region}
\end{equation}
\begin{wrapfigure}{r}{0.5\linewidth}
  \centering
  \includegraphics[width=1\linewidth]{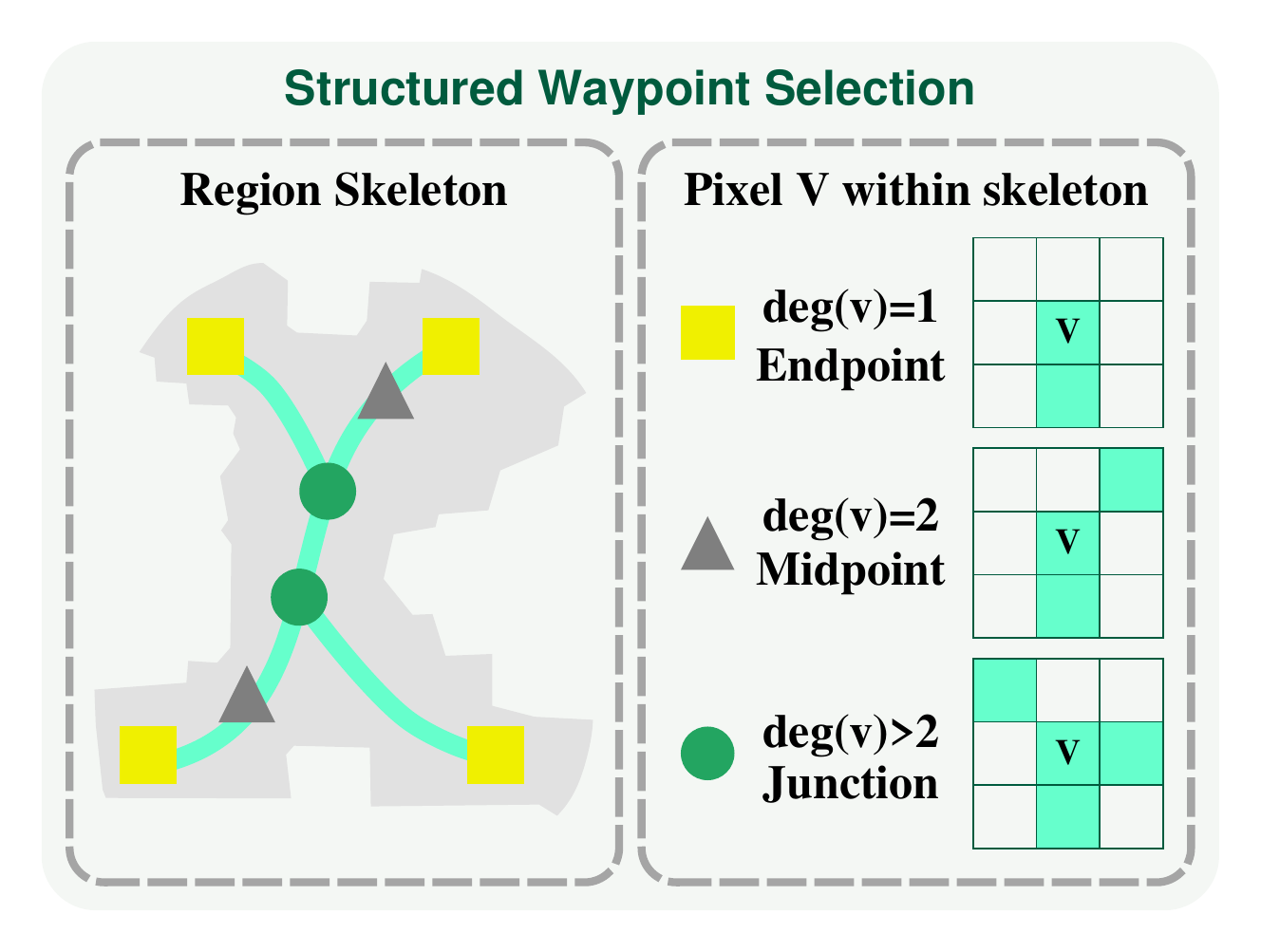}
  \caption{\textbf{Structured waypoint selection based on skeleton.} We categorize skeleton nodes by their degree and select only degree-1 endpoints as candidate waypoints.}
  \label{fig:select}
\end{wrapfigure}
where $\delta_h$ is the height threshold, $r$ is the local planning radius, and $\Pi(\cdot)$ denotes orthographic projection onto the horizontal plane. $\Omega_t$ defines the local traversable area around the agent.

\textbf{Topological Skeleton Extraction.}
To introduce structural priors into the local decision space, we abstract the raw traversable region $\Omega_t$ into a topological skeleton $\mathcal{S}_t$ using morphological thinning~\cite{lee1994building}:
\begin{equation}
\mathcal{S}_t = {Skeletonize}(\Omega_t),
\label{eq:skeleton}
\end{equation}
where $\mathcal{S}_t \subset \Omega_t$ approximates the center axis of free space. Rather than interpreting navigable space as a dense, unstructured area, the skeleton captures its underlying topology by tracing the central lines of movement through open regions. 
This is analogous to how humans often form mental maps of environments based on key corridors, intersections, and doorways, rather than memorizing complete spatial coverage. By reasoning over this abstract structure, the agent can better plan paths that respect the environment's layout and reduce unnecessary action noise.

\textbf{Structured Waypoint Selection.}
We model the skeleton $\mathcal{S}_t$ as an undirected graph $\mathcal{G}_t = (\mathcal{V}_t, \mathcal{E}_t)$, where nodes $\mathcal{V}_t$ correspond to skeleton pixels and edges $\mathcal{E}_t$ reflect 8-connected adjacency (including horizontal, vertical, and diagonal neighbors in the 2D grid). 
To reduce redundancy and focus on directionally meaningful locations, we select a sparse set of endpoints from the graph as candidate waypoints. Specifically, we retain only nodes with degree 1:
\begin{equation}
\mathcal{W}_t = \left\{ v_i \in \mathcal{V}_t \;\middle|\; \deg(v_i) = 1 \right\},
\label{eq:waypoints}
\end{equation}
which correspond to the outermost reachable points on the local skeleton, as shown in Fig.~\ref{fig:select}. These endpoints naturally capture the agent's forward navigability and latent path divergence. Although we do not explicitly select junctions, their future branches are represented by the endpoints along each subpath.

Each selected waypoint $w_i \in \mathcal{W}_t$ is projected back into 3D. We input the RGB views $\mathbf{I}_t$ into a VLM to extract semantic information, focusing on visual content in the direction of each waypoint. For each $w_i$, the VLM outputs a textual description $\mathcal{D}_i$ of the corresponding direction.
The final decision space is defined as a set of paired spatial-semantic candidates:
\begin{equation}
\mathcal{A}_t = \left\{ (w_i, \mathcal{D}_i) \;\middle|\; w_i \in \mathcal{W}_t,\ \mathcal{D}_i = \text{VLM}(\textbf{I}_t, w_i) \right\},
\label{eq:decision_space}
\end{equation}
which couples structural layout with perceptual grounding. This layout-constrained action space forms part of the input for the LLM.

\subsection{Task-Aligned Feedback Regulation}  
\label{sec:3.3}

To ensure instruction-aligned behavior over long-horizon trajectories, we introduce a feedback regulation mechanism that dynamically adjusts the agent's decision space based on recent observations and subtask progression. This module plays a central role in our principle of \textit{Instruction-Aligned Decision Space Optimization}, allowing the agent to refine its actions not only based on spatial layout but also on semantic alignment with the instruction over time.

\textbf{Visual Feedback Generation.}
Through LLM reasoning, the agent selects a waypoint $w_t^*$ from the decision space $\mathcal{A}_t$, resulting in an action $a_t$. After executing $a_t$, it receives the next observation $\textbf{O}_{t+1} = (\textbf{I}_{t+1}, \textbf{D}_{t+1})$.
To assess whether the agent has made progress toward the current subtask $\mathcal{T}_t$ (derived from instruction $L$), we input the observation pair $(\textbf{O}_t, \textbf{O}_{t+1})$ and the subtask $\mathcal{T}_t$ into the VLM to generate a feedback signal:
\begin{equation}
f_{t+1} = \text{VLM}(\textbf{O}_t, \textbf{O}_{t+1}, \mathcal{T}_t).
\label{eq:feedback}
\end{equation}
where $f_{t+1}$ is a textual reflection describing the change in scene with respect to $\mathcal{T}_t$---e.g., “partially entered the bedroom” or “moved away from the target.” The feedback offers fine-grained progress estimation at each step, allowing the agent to detect incremental advances or errors.

\textbf{Feedback-Guided Action Selection.}
To determine the next action, the agent reasons over the updated structured decision space $\mathcal{A}_{t+1}$ (similarly defined in Eq.~\eqref{eq:decision_space}), the full instruction $L$, and the generated feedback $f_{t+1}$. The LLM outputs the next action via:
\begin{equation}
w_{t+1}^* = \text{LLM}(\mathcal{A}_{t+1}, f_{t+1}, L),
\label{eq:llm_decision}
\end{equation}
\begin{equation}
a_{t+1} = Action(w_{t+1}^*).
\label{eq:action}
\end{equation}
This loop forms a closed decision-feedback cycle, in which action selection is continuously guided by semantic progress monitoring. As the agent moves through the environment, each observation is not only encoded visually but interpreted in light of task intent, enabling corrective adjustments and reducing cumulative drift. In this way, the decision space is adaptively regulated by execution context, maintaining semantic coherence with the instruction over time.
By iterating this closed-loop process for $T$ steps, the agent produces an action sequence $\{a_1, a_2, \dots, a_T\}$ that successfully navigates toward the instruction-aligned target.

\section{Experiments} \label{Experiments}


\subsection{Experimental Setup}

\textbf{R2R-CE Dataset.} 
We conduct experiments on the R2R-CE dataset, which extends the Room-to-Room (R2R) benchmark for visual language navigation (VLN)~\cite{anderson2018vision, krantz2020beyond}. This dataset consists of natural language instructions paired with navigation trajectories in realistic 3D indoor environments, derived from the Matterport3D dataset~\cite{chang2017matterport3d}. We follow the settings of OpenNav~\cite{qiao2024open}, conducting tests on 100 randomly selected episodes from the dataset. In these experiments, we leverage both VLM and LLM to perform zero-shot navigation. Our goal is to fully leverage the generalization and reasoning capabilities of these pretrained models, enabling them to adapt to the current task without any additional training.

\textbf{RxR-CE Dataset.} 
We also use the RxR-CE dataset, which extends the Room-Across-Room (RxR) benchmark with similar challenging conditions~\cite{ku2020room, krantz2020beyond}. RxR features longer and more diverse instructions across multiple languages and emphasizes global navigation capabilities. The CE variant (Continuously Evolving) simulates viewpoint changes and environmental variations, making it suitable for evaluating the robustness and generalization of navigation agents under distribution shifts.

\textbf{Evaluation metrics.} 
We evaluate navigation performance using standard metrics. Navigation Error (NE) measures the shortest-path distance between the agent's final position and the goal. Success Rate (SR) is the percentage of episodes where the agent stops within 3 meters of the goal. Success weighted by Path Length (SPL) balances success and path efficiency~\cite{anderson2018vision}. Normalized Dynamic Time Warping (NDTW) reflects the similarity between the predicted and reference trajectories~\cite{ilharco2019general}. Oracle Success Rate (OSR) indicates the best possible success assuming the agent stops optimally along its path~\cite{krantz2020beyond}. Trajectory Length (TL) records the average length of agent trajectories. Soft-DTW (SDTW) is a relaxed version of DTW that tolerates slight deviations in trajectory matching~\cite{cuturi2017soft}.

\textbf{Implementation details.} 
All experiments are conducted in simulated VLN-CE environments. At each step, the agent receives an RGB-D observation, where the RGB input is resized to $244 \times 244 \times 3$ and the depth map to $256 \times 256$. Structured waypoints are generated by extracting skeletons from depth without relying on any pretrained waypoint predictor. For perception and feedback generation, we use {Qwen-VL-Max} as the Vision-Language Model (VLM). The action selection process is guided by {GPT-4o}, which reasons over the instruction, structured perception, and feedback to choose the next waypoint. Our VLM and LLM are accessed via API rather than deployed locally; for local deployment using open-source models, please refer to Open-Nav~\cite{qiao2024open}.

\subsection{Main Results on Zero-Shot VLN-CE}
\label{sec:4.2}
\begin{table}[t]
    \setlength{\tabcolsep}{5pt}
    \renewcommand{\arraystretch}{0.9} 
    \caption{\textbf{Performance comparison on the R2R-CE dataset.} The dash "–" indicates that the corresponding metric was not reported in the original work. \textbf{Bold} indicates the best result. We report the relative percentage change of our method compared to the previous SOTA in parentheses. \textcolor{red}{Red} denotes improvement, while \textcolor{green}{Green} indicates degradation.}
    \label{tab:r2r-ce}
    \centering
    \begin{tabular}{lccccc}
        \toprule
        Method  & NE$\downarrow$ & NDTW$\uparrow$ & OSR$\uparrow$ & SR$\uparrow$ & SPL$\uparrow$ \\
        \midrule
        \multicolumn{6}{c}{\textbf{Supervised Learning}} \\
        \midrule
        CMA~\cite{krantz2020beyond}  & 6.92 & 50.77 & 45 & 37 & 32.17 \\
        RecBERT~\cite{wu2024recbert}  & 5.8 & 54.81 & 57 & 48 & 43.22 \\
        BEVBert~\cite{an2022bevbert}  & 5.13 & 61.40 & 64 & 60 & 53.41 \\
        ETPNav~\cite{an2024etpnav}  & 5.15 & 61.15 & 58 & 52 & 52.18 \\
        HNR~\cite{wang2024lookahead}  & 4.42 & - & 67 & 61 & 51 \\
        \midrule
        \multicolumn{6}{c}{\textbf{Zero-Shot}} \\
        \midrule
        Random~\cite{qiao2024open} & 8.63 & 34.08 & 12 & 2 & 1.50 \\
        LXMERT~\cite{krantz2020beyond}  & 10.48 & 18.73 & 22 & 2 & 1.87 \\
        DiscussNav-GPT4~\cite{long2024discuss}  & 7.77 & 42.87 & 15 & 11 & 10.51 \\
        Open-Nav-Llama3.1~\cite{qiao2024open}  & 7.25 & 44.99 & 23 & 16 & 12.90 \\
        Open-Nav-GPT4~\cite{qiao2024open}  & \textbf{6.70} & 45.79 & 23 & 19 & 16.10 \\
        SmartWay~\cite{shi2025smartway}  & 7.01 & - & \textbf{51} & 29 & 22.46 \\
        \textbf{Ours}  & 6.91\textcolor{green}{(3.1\%)} & \textbf{51.8}\textcolor{red}{(13.2\%)} & 39\textcolor{green}{(23.5\%)} & \textbf{35}\textcolor{red}{(20.6\%)} & \textbf{30.30}\textcolor{red}{(34.9\%)} \\
        \bottomrule
    \end{tabular}
\end{table}

\begin{table}[t]
    \setlength{\tabcolsep}{5pt}
    \renewcommand{\arraystretch}{0.9} 
    \caption{\textbf{Performance comparison on the RXR-CE dataset.} The dash "-" indicates that the corresponding metric was not reported in the original work. \textbf{Bold} indicates the best result. We report the relative percentage change of our method compared to the previous SOTA in parentheses. \textcolor{red}{Red} denotes improvement, while \textcolor{green}{Green} indicates degradation.}
    \label{tab:rxr-ce}
    \centering
    \begin{tabular}{lccccc}
        \toprule
        Method    & NE$\downarrow$ & NDTW$\uparrow$ & SDTW$\uparrow$ & SR$\uparrow$ & SPL$\uparrow$ \\ 
        \midrule
        \multicolumn{6}{c}{\textbf{Supervised Learning}} \\
        \midrule
        LAW~\cite{raychaudhuri2021language}       & 11.04 & 37.0 & 8.0 & 10.0 & 9.0 \\
        VLN\circlearrow BERT~\cite{hong2020recurrent} & 8.98 & 46.7 & - & 27.1 & 23.7 \\
        GridMM ~\cite{wang2023gridmm}  & 8.42 & 48.2 & 33.7 & 36.3 & 30.1 \\
        ETPNav~\cite{an2024etpnav}   & 5.64 & 61.9 & 45.3 & 54.8 & 44.9 \\
        WS-MGMap~\cite{chen2022weakly} & 9.83 & - & - & 15.0 & 12.1 \\
        HNR~\cite{wang2024lookahead} & 5.51 & 63.56 & 47.24 & 56.39 & 46.73 \\ 
        \midrule
        \multicolumn{6}{c}{\textbf{Zero-Shot}} \\
        \midrule
        A\textsuperscript{2}Nav~\cite{chen20232} & - & - & - & 16.8 & 6.3 \\ 
        CA-Nav~\cite{chen2024constraint}  & \textbf{10.37} & 13.5 & 5.0 & 19.0 & 6.0 \\
        \textbf{Ours} & 11.19\textcolor{green}{(7.9\%)} & \textbf{30.1}\textcolor{red}{(122.9\%)} & \textbf{8.9}\textcolor{red}{(78.0\%)} & \textbf{21.2}\textcolor{red}{(11.5\%)} & \textbf{9.6}\textcolor{red}{(52.3\%)} \\
        \bottomrule
    \end{tabular}
\end{table}

We evaluate STRIDER on two standard zero-shot VLN-CE benchmarks: R2R-CE (Tab.~\ref{tab:r2r-ce}) and RxR-CE (Tab.~\ref{tab:rxr-ce}). Across both datasets, STRIDER consistently outperforms prior zero-shot methods on key metrics such as SPL, SR, and NDTW, indicating more reliable goal completion and higher trajectory fidelity. While supervised methods benefit from task-specific training, STRIDER remains competitive despite operating without fine-tuning or environment-specific adaptation.

On R2R-CE, STRIDER achieves substantial improvements in SPL and NDTW over all prior zero-shot models, driven by two key design factors. The Structured Waypoint Generator constrains the agent's behavior to layout-consistent paths, reducing detours and spatial drift, while the Task-Alignment Regulator provides real-time feedback to maintain semantic alignment and correct deviations. This combination enables STRIDER to balance spatial feasibility and instruction adherence, leading to gains across both path-quality and goal-completion metrics. On the more diverse RxR-CE benchmark, STRIDER continues to outperform other zero-shot methods, though with narrower margins—likely due to RxR's higher linguistic and geographic variability. Even in such settings, STRIDER's structured decision space offers a strong prior that compensates for ambiguity, while the feedback loop supports adaptive planning. Together, these elements demonstrate the robustness and generality of decision space optimization under zero-shot conditions.

It is worth mentioning that STRIDER does not achieve the lowest Navigation Error (NE), which is expected given its emphasis on instruction alignment and structural feasibility over exact endpoint proximity. The agent may stop at semantically appropriate locations slightly offset from the goal, reflecting a preference for coherent, interpretable paths over shortcut-based precision---an acceptable tradeoff in instruction-guided navigation tasks.

\begin{table}[t]
\renewcommand{\arraystretch}{0.9} 
\caption{\textbf{Effect of Structured Waypoint Generator (SWG) under different node degree configurations.} \textbf{Bold} indicates the best result. We report the relative percentage change compared to the baseline in parentheses. \textcolor{red}{Red} denotes improvement, while \textcolor{green}{Green} indicates degradation. The gray-shaded row denotes the primary experimental configuration used in our main results.}
\label{tab:swg-ablation}
\centering
\begin{tabular}{cc|ccccc}
\toprule
SWG & Node Degree & NE$\downarrow$ & NDTW$\uparrow$ & OSR$\uparrow$ & SR$\uparrow$ & SPL$\uparrow$ \\
\midrule
 \XSolidBrush & \XSolidBrush  & 7.19 & 48.78 & 29 & 24 & 21.07 \\
 \rowcolor{gray!20}
 \Checkmark & 1  & 6.91\textcolor{red}{(3.9\%)} & \textbf{51.87}\textcolor{red}{(6.3\%)} & \textbf{39}\textcolor{red}{(34.5\%)} & \textbf{35}\textcolor{red}{(45.8\%)} & \textbf{30.30}\textcolor{red}{(43.8\%)} \\
 \Checkmark & > 2  & 7.34\textcolor{green}{(2.1\%)} & 49.88\textcolor{red}{(2.3\%)} & 33\textcolor{red}{(13.8\%)} & 28\textcolor{red}{(16.7\%)} & 25.02\textcolor{red}{(18.7\%)} \\
 \Checkmark & $\neq$ 2  & \textbf{6.83}\textcolor{red}{(5.0\%)} & 51.12\textcolor{red}{(4.8\%)} & 38\textcolor{red}{(31.0\%)} & 33\textcolor{red}{(37.5\%)} & 29.21\textcolor{red}{(38.6\%)} \\
\bottomrule
\end{tabular}
\end{table}

\begin{figure*}[t]
  \centering
  \includegraphics[width=1\linewidth]{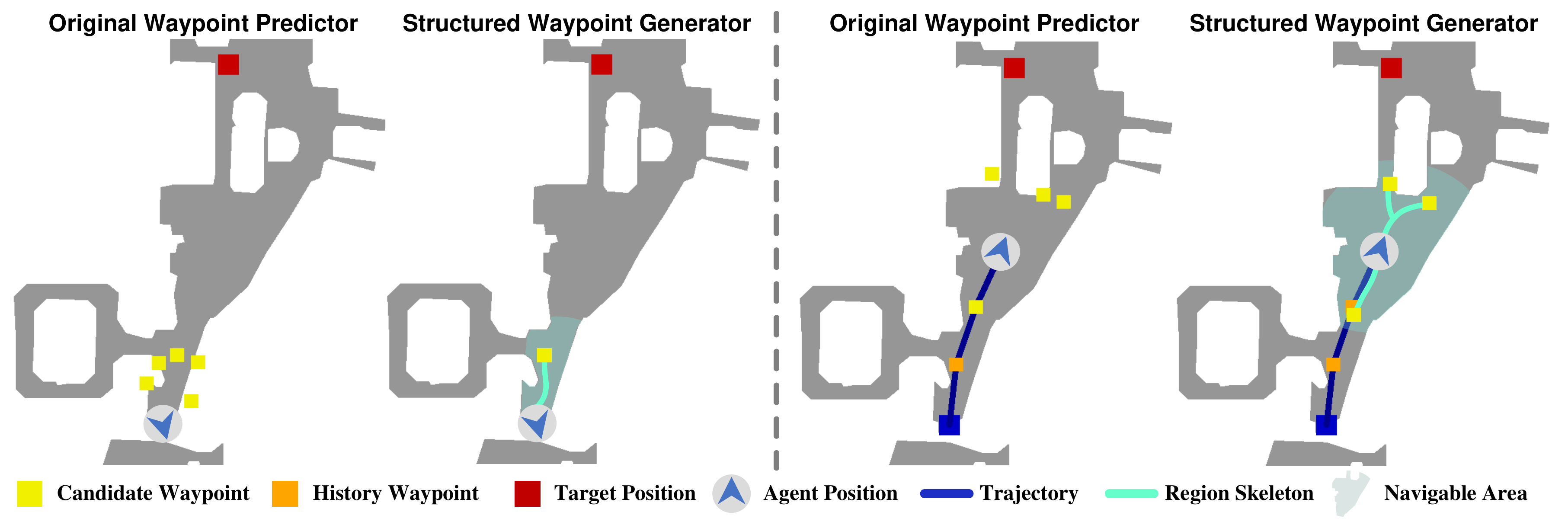}
  \caption{\textbf{Comparison between original waypoint predictor and Structured Waypoint Generator.} Our Structured Waypoint Generator extracts layout-consistent waypoints that align with the environment's topology, resulting in trajectories that are more goal-directed and spatially coherent.}
  \label{fig:wp}
\end{figure*}

\subsection{Ablation Study}
\label{sec:4.3}

We conduct ablation experiments to analyze the impact of STRIDER's two core modules: the Structured Waypoint Generator and the Task-Alignment Regulator. For each module, we present both quantitative comparisons and qualitative visualizations to highlight their effects on navigation performance.

\subsubsection{Effect of Structured Waypoint Generator}

We evaluate different configurations of the Structured Waypoint Generator to understand how layout-aware guidance affects agent behavior. As shown in Tab.~\ref{tab:swg-ablation}, we consider two key design choices: (1) whether to use the Structured Waypoint Generator, and (2) how to guide waypoint selection based on skeleton connectivity---e.g., using only endpoint nodes (deg($v_i$)=1), higher-degree junctions (deg($v_i$)>2) or both. We use the original waypoint predictor trained on the R2R dataset as the baseline for STRIDER without the Structured Waypoint Generator. We present the visualization comparison between the original waypoint predictor and the Structured Waypoint Generator in Fig.~\ref{fig:wp}.

Using degree-1 nodes (endpoints) yields the best overall performance across all metrics, as these points typically lie at corridor tips or face the goal direction, providing clear guidance with minimal ambiguity. In contrast, using only high-degree nodes ($>$2), such as junctions, leads to degraded performance with higher NE (7.34) and lower SR/SPL, likely due to increased behavioral uncertainty. Combining non-2-degree nodes (i.e., $\neq$2) provides a good compromise, incorporating both endpoints and informative junctions to achieve strong performance (NE: 6.83, SPL: 29.21) at the cost of a slightly longer trajectory. Based on these results, we adopt the degree-1 configuration as the default in our main experiments.

\subsubsection{Effect of Task-Alignment Regulator}

\begin{table}[t]
\renewcommand{\arraystretch}{0.9} 
\caption{\textbf{Effect of the Task-Alignment Regulator (TAR) on navigation performance.} \textbf{Bold} indicates the best result. We report the relative percentage change compared to the baseline in parentheses. \textcolor{red}{Red} denotes improvement, while \textcolor{green}{Green} indicates degradation. The gray-shaded row denotes the primary experimental configuration used in our main results.}
\label{tab:TAR-ablation}
\centering
\begin{tabular}{c|ccccc}
\toprule
TAR & NE$\downarrow$ & NDTW$\uparrow$ & OSR$\uparrow$ & SR$\uparrow$ & SPL$\uparrow$ \\
\midrule
 \XSolidBrush & \textbf{6.77} & 51.06 & \textbf{42} & 29 & 26.11 \\
 \rowcolor{gray!20}
 \Checkmark  & 6.91\textcolor{green}{(2.1\%)} & \textbf{51.87}\textcolor{red}{(1.6\%)} & 39\textcolor{green}{(7.1\%)} & \textbf{35}\textcolor{red}{(20.7\%)} & \textbf{30.30}\textcolor{red}{(16.0\%)} \\
\bottomrule
\end{tabular}
\end{table}

\begin{figure*}[t]
  \centering
  \includegraphics[width=1\linewidth]{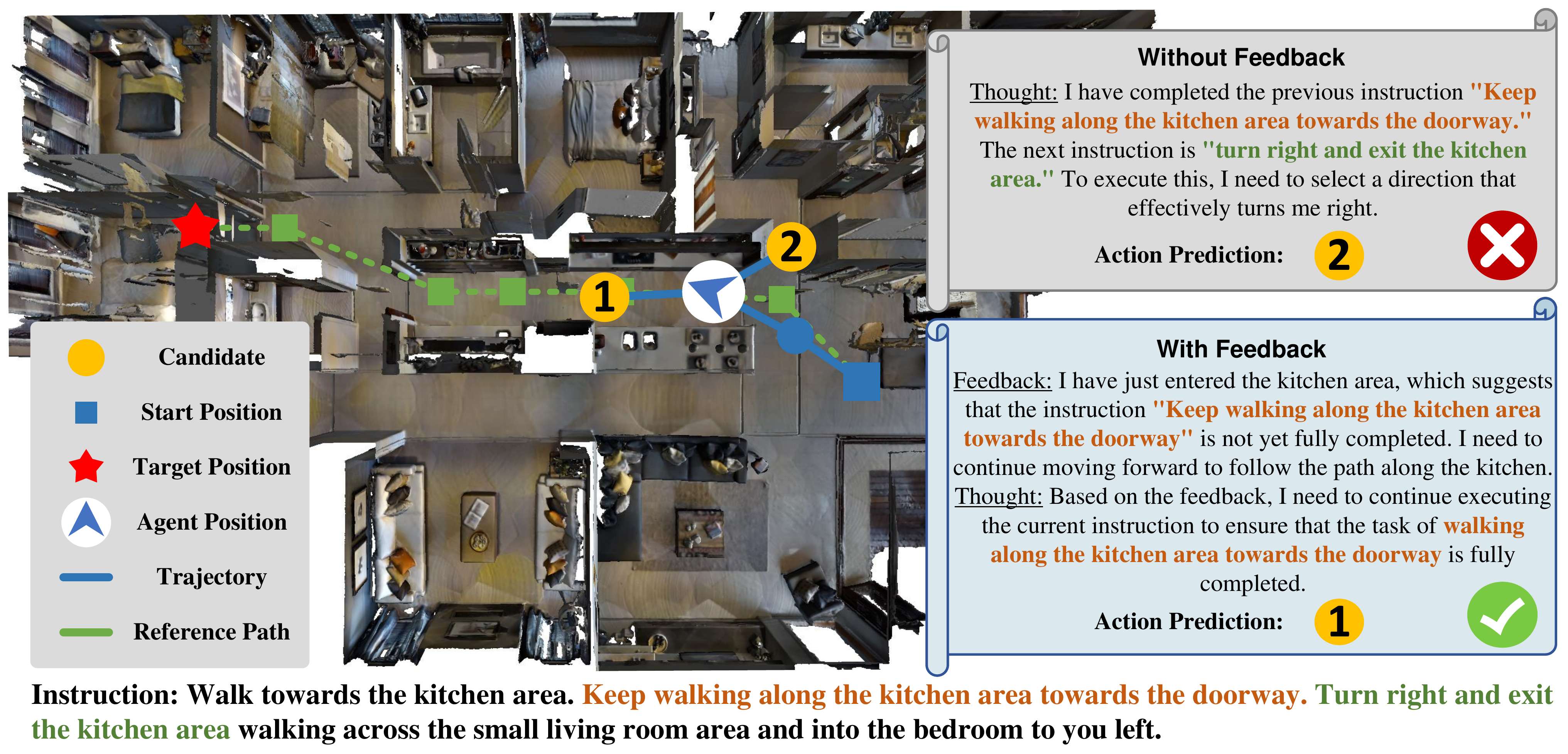}
  \caption{\textbf{Comparison of agent behavior under no-feedback and feedback-driven execution strategies.} Without feedback, the agent prematurely infers task completion, resulting in an incorrect action (Action 2). With feedback, the agent leverages the intermediate state to refine its understanding, yielding a more semantically consistent action (Action 1) aligned with the instruction.}
  \label{fig:vis1}
\end{figure*}

As shown in Tab.~\ref{tab:TAR-ablation}, enabling the Task-Alignment Regulator (TAR) leads to consistent improvements across SPL, SR, and NDTW, indicating that feedback-driven behavior adjustment helps the agent maintain semantic alignment and recover from drift during long-horizon navigation. While NE slightly increases (6.77→6.91), this reflects a more conservative execution pattern, where the agent prioritizes instruction fidelity and avoids premature termination, ultimately leading to higher task success and trajectory consistency. We illustrate the impact of the Task-Alignment Regulator on navigation behavior in Fig.~\ref{fig:vis1}. Overall, TAR enhances instruction fidelity without compromising spatial plausibility and is adopted in our main model configuration.

\subsubsection{Effect of Model-Agnostic Design}  
STRIDER is not tied to any single model and can operate effectively across components of similar capabilities.
To verify this, we conduct experiments using various VLMs of different sizes and providers.
As shown in Tab.~\ref{tab:vlm-ablation}, GPT-4o achieves the best performance in terms of NE (6.75) and SR (36). However, STRIDER also shows competitive results using similar models, such as Qwen-VL-Max and Qwen2.5-VL-72B, which perform strongly in NDTW, OSR and SR, while still maintaining an overall solid performance. Additionally, STRIDER continues to deliver good results with smaller models like Qwen2.5-VL-32B and Qwen2.5-VL-7B.
The consistent performance across different models suggests that STRIDER's design amplifies the strengths of the underlying model without relying on a specific model. 

For our primary experiments, we selected Qwen-VL-Max. This choice highlights STRIDER's model-agnostic design, which enables our method to work with any similar model and ensures that its results are driven by the strength of our approach, rather than reliance on a specific foundation model.

\begin{table}[t]
\setlength{\tabcolsep}{9pt}
\renewcommand{\arraystretch}{0.95} 
\caption{\textbf{Ablation on different VLMs.} We test our method using various VLMs of different sizes and capabilities. \textbf{Bold} indicates the best result. The gray-shaded row denotes the primary experimental configuration used in our main results.}
\label{tab:vlm-ablation}
\centering
\begin{tabular}{c|cccccc}
\toprule
VLM & TL & NE$\downarrow$ & NDTW$\uparrow$ & OSR$\uparrow$ & SR$\uparrow$ & SPL$\uparrow$ \\
\midrule
\rowcolor{gray!20}
Qwen-VL-Max      & 8.13  & 6.91  & 51.87 & \textbf{39} & 35 & 30.30 \\
Qwen2.5-VL-72B   & 8.30  & 6.78  & 51.99 & 39 & 34 & 29.07 \\
Qwen2.5-VL-32B   & 8.56  & 7.12  & 48.02 & 33 & 28 & 24.20 \\
Qwen2.5-VL-7B    & 8.92  & 7.46  & 46.35 & 29 & 24 & 21.12 \\
GPT-4o           & 8.01  & \textbf{6.75}  & 50.12 & 39 & \textbf{36} & \textbf{31.37} \\
Gemini-2.5-Pro   & 8.34  & 6.92  & 51.35 & 37 & 34 & 29.85 \\
Gemini-2.5-Flash & 7.68  & 7.08  & 49.87 & 34 & 29 & 25.30 \\
Claude-3.5       & 7.81  & 6.86  & \textbf{52.10} & 36 & 33 & 29.40 \\
Claude-4         & 8.22  & 7.14  & 45.25 & 31 & 29 & 26.10 \\
\bottomrule
\end{tabular}
\end{table}

\subsubsection{Applying SWG to Fine-Tuned Models}

We further assess the effectiveness of Structured Waypoint Guidance (SWG) in fine-tuned models. Specifically, we incorporate our SWG module into the BEVBert model by replacing its original waypoint prediction mechanism. This modification leads to improvements across all key evaluation metrics, as shown in Table~\ref{tab:finetune-swg}.

Even in fine-tuned settings, SWG's ability to integrate environmental structure as a strong prior helps compensate for uncertainties in the navigation task. This further reinforces the versatility and effectiveness of SWG in different contexts, demonstrating that structured priors can be seamlessly integrated into existing models, thereby improving their robustness and reliability across a range of navigation tasks.

\begin{table}[t]
\renewcommand{\arraystretch}{0.9} 
\centering
\caption{\textbf{Applying SWG to BEVBert.} We compare the vanilla BEVBert with the one applying our SWG. \textbf{Bold} indicates the best result. We report the relative percentage change compared to the baseline in parentheses. \textcolor{red}{Red} denotes improvement.}
\label{tab:finetune-swg}
\begin{tabular}{c|cccc}
\toprule
Method & NE$\downarrow$ & OSR$\uparrow$ & SR$\uparrow$ & SPL$\uparrow$ \\
\midrule
BEVBert & 4.57 & 67 & 59 & 50 \\
\textbf{BEVBert w/ SWG} & \textbf{4.37}\textcolor{red}{(4.3\%)} & \textbf{70}\textcolor{red}{(4.4\%)} & \textbf{61}\textcolor{red}{(3.3\%)} & \textbf{53}\textcolor{red}{(6.0\%)} \\
\bottomrule
\end{tabular}
\end{table}

\section{Conclusion}

We presented STRIDER, a zero-shot VLN-CE framework that optimizes agent behavior through Instruction-Aligned Structural Decision Space Optimization. By combining a Structured Waypoint Generator with a Task-Alignment Regulator, STRIDER enables agents to navigate in complex, unseen environments in a manner that is both structurally coherent and semantically faithful to natural language instructions. Extensive experiments on VLN-CE benchmarks demonstrate that our approach outperforms strong zero-shot baselines across multiple metrics, with ablations confirming the complementary contributions of structural planning and feedback-driven regulation. This work highlights the importance of structuring and modulating the decision space for long-horizon instruction following and opens up future directions in integrating richer spatial priors, adaptive subgoal modeling, and instruction-aware exploration strategies.

\section{Acknowledgement}

This work was supported by the National Natural Science Foundation of China under Grant 62293543, Grant 62322605.

	{\small
		\bibliographystyle{plain}
		\bibliography{ref}
	}

\newpage

\appendix

\section*{Supplemental Material}

The supplemental material is organized as follows:
\begin{itemize}
    \item \ref{sec:implementation} \quad \textbf{Implementation Details}: implementation details of STRIDER, including the system hyperparameters, prompting setup, and deployment configurations.
    \item \ref{sec:robustness} \quad \textbf{Robustness Analysis}: robustness analysis under two settings: perceptual degradation and mid-trajectory perturbation.
    \item \ref{sec:visualization} \quad \textbf{Visualization Results}: qualitative visualization results, including a step-by-step case study, additional episodes, and failure cases.
    \item \ref{sec:limitation} \quad \textbf{Limitations and Future Work}: discussion of current limitations related to perception settings and foundation model dependencies, along with future directions.
    \item \ref{sec:impact} \quad \textbf{Broader Impact}: broader impact analysis concerning the societal implications of deploying large-scale VLM- and LLM-based navigation systems.
\end{itemize}

These sections provide further insights into our system design, experimental robustness, and qualitative behaviors, along with reflections on limitations and potential societal implications.

\section{Implementation Details}
\label{sec:implementation}

In this section, we provide comprehensive implementation details of our proposed STRIDER. We begin by outlining the overall system pipeline and deployment setup, including the perception configuration, structured waypoint generation, task-alignment mechanism, and LLM-based planning loop. We then describe the design of our prompting templates for the Vision-Language Model (VLM), which play a central role in both environmental perception and progress feedback.

\subsection{System Pipeline and Deployment Settings}

All experiments were conducted in a simulated VLN-CE environment using the Matterport3D simulator. We followed the standard evaluation protocols defined by the R2R-CE and RxR-CE benchmarks.

\textbf{Perception Setup.} 
At each timestep $t$, the agent receives a 360\textdegree{} RGB-D panoramic observation consisting of 12 RGB-D frames spaced at 30\textdegree{} intervals. All frames are captured at a fixed horizontal elevation, without looking above or below eye level. The RGB images are resized to $244 \times 244 \times 3$, and the depth maps are resized to $256 \times 256$. To obtain a holistic understanding of the surrounding environment, we concatenate the 12 RGB frames into a single $3 \times 4$ grid image and input it into the Vision-Language Model (VLM) to generate a global scene description. This allows the agent to build a comprehensive understanding of the overall context. Additionally, for each candidate waypoint, we feed the corresponding directional RGB view into the model to obtain a localized visual description. We adopt Qwen-VL-Max (accessed via API) as the default VLM for all perception and feedback-related tasks in our system. 

\textbf{Structured Waypoint Generation.}  
We extract spatially structured waypoints by processing the local point cloud centered on the agent. Points with vertical height below $\delta_h=-1.0 \text{m}$ are retained to approximate surrounding static structures. These points are projected onto the \(x\text{-}z\) plane to form a 2D occupancy map.
To obtain a clean and geometrically consistent traversable area, we apply contour filling followed by Gaussian smoothing (kernel size \(75 \times 75\)), which helps simplify irregular boundaries and suppress small artifacts, thereby reducing unnecessary branches in the subsequent skeleton. The result is then binarized and filtered to retain only the largest connected region.
We extract a topological skeleton via morphological thinning and identify endpoints as pixels with degree = 1 in the 8-connected neighborhood, corresponding to path-extremities in the navigable space. These endpoints are merged within a 10-pixel radius to reduce redundancy, and converted back to world coordinates. Points within 1.0 meter of the agent are excluded to avoid overly local actions. The remaining candidates, expressed in relative polar coordinates (distance and direction), constitute the final structured waypoint set.

\textbf{Task-Alignment Regulator.}  
After each action, the agent evaluates whether its behavior has advanced toward the current subgoal by comparing visual observations before and after the movement. Specifically, we select a single RGB frame aligned with the selected waypoint direction from the pre-action panoramic observation \(O_t\), and the front-facing RGB frame from the post-action observation \(O_{t+1}\). These two views are concatenated side-by-side into a single image, forming a minimal visual trajectory context for comparison.
This composite image, along with the current subtask parsed from the instruction (e.g., ``go into the bedroom''), is fed into the VLM. The model generates a natural language description indicating the perceived semantic change. This feedback is then passed to the LLM planner to guide the next decision step, effectively closing the perception-action loop with dynamic task progress estimation.

\begin{figure*}[t]
  \centering
  \includegraphics[width=1\linewidth]{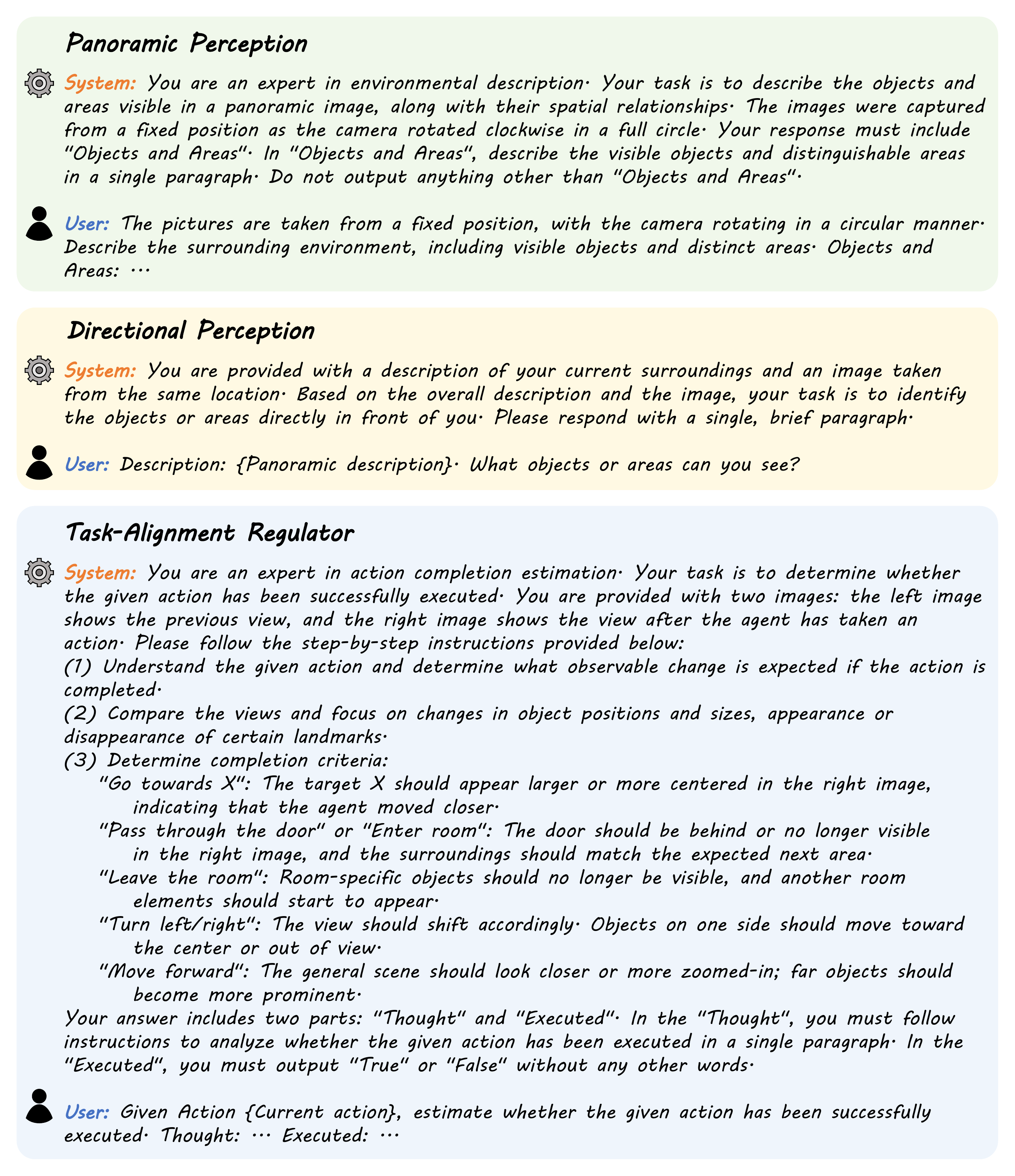}
  \caption{\textbf{Overview of the designed prompt templates.} Panoramic Perception prompts guide the VLM to generate global semantic summaries from 360° panoramic views. Directional Perception prompts provide localized visual descriptions for each candidate waypoint. Task-Alignment Regulator prompts enable the system to assess action execution status by comparing visual observations before and after navigation steps.}
  \label{fig:prompt}
\end{figure*}

\textbf{LLM-Based Planning.}  
We use GPT-4o (accessed via API) as the decision-maker. At each step, GPT-4o receives the instruction \(L\), the current decision space \(\mathcal{A}_t\), and the feedback \(f_t\) to decide on the next waypoint \(w^*_t \in W_t\). After making a decision, the LLM also generates a short explanation of its reasoning, such as why a particular direction is selected, and maintains a concise history of previous actions and their corresponding explanations.

\textbf{Stopping Strategy.}  
STRIDER does not rely on the model generating an explicit {STOP} signal. Instead, the instruction is first decomposed into a series of subtasks by the LLM, with the maximum number of execution steps set to match the number of subtasks, ensuring a hard minimum of 6 steps to guarantee task completion. Once the maximum step is reached, the system halts. This approach eliminates reliance on potentially unreliable stop predictions from the model and guarantees stable behavior across tasks.

\textbf{Hardware.}  
We conduct all inference experiments on a single NVIDIA RTX 4090 GPU. Since both the VLM and the LLM are accessed through APIs rather than deployed locally, and the original trained waypoint predictor is discarded in favor of structured skeleton-based planning, the overall GPU memory footprint of our system is minimal. This makes our pipeline lightweight and feasible to run. For users interested in local deployment of the LLM, we recommend referring to the hardware setup used in the Open-Nav as a guideline.

\subsection{Prompting Templates}

We design a set of prompt templates for the VLM tailored to two key components: perception and feedback generation. These prompts are crafted to guide the VLM in producing meaningful scene descriptions, localized object information, and action-oriented feedback.

Specifically, our perception prompts aim to extract global understanding from panoramic visual inputs, as well as detailed descriptions from directional candidate views. Meanwhile, our feedback prompts are primarily designed to guide the VLM in assessing task completion status and generating textual outputs accordingly. We present the prompt templates in Fig.~\ref{fig:prompt}.

\section{Robustness Analysis}
\label{sec:robustness}

In this section, we evaluate the robustness of our STRIDER under two types of realistic challenges: sensory sparsity and mid-execution perturbations.

\subsection{Robustness to Perceptual Degradation}

To validate the model's robustness to sparser perceptual inputs, we deliberately reduce the robot's accessible RGB-D observations from 12 to 6, increasing the angular interval from $30^\circ$ to $60^\circ$. This setting directly challenges the model’s ability to maintain spatial awareness and semantic consistency with less comprehensive visual coverage. We take Open-Nav-GPT4 as a baseline for comparison to assess the relative robustness of different models under the same degradation condition.

As shown in Table~\ref{tab:robust-perception}, both Open-Nav-GPT4 and STRIDER experience significant performance drops under degraded perception, yet the degree of robustness differs markedly between them. For {Open-Nav-GPT4}, reducing the number of views results in a 20.45\% increase in navigation error (NE), along with substantial decreases in success-related metrics: success rate (SR) drops by 47.37\%, and success weighted by path length (SPL) drops by 46.73\%. This indicates that the model's path planning and decision confidence are heavily reliant on high-resolution panoramic context. Furthermore, the decrease in normalized dynamic time warping (NDTW) by 12.02\% suggests degraded path adherence and reduced semantic trajectory alignment.

In contrast, {STRIDER} demonstrates relatively greater robustness. While it also suffers from performance degradation, the increases in NE (+13.33\%) and decreases in NDTW (–7.49\%) and SR (–34.29\%) are more moderate. Notably, SPL drops by 32.47\%, which, though significant, is still considerably less than the degradation observed in Open-Nav-GPT4. This suggests that STRIDER benefits from either more efficient path reasoning or a less perception-heavy reliance on global context.

Overall, the analysis reveals that while both models are sensitive to perceptual sparsity, STRIDER retains higher performance under degraded input, potentially due to more robust spatial priors or stronger local visual grounding. These findings highlight the importance of designing VLM-based agents that are resilient to variations in sensory input density, especially in real-world deployment scenarios where perceptual completeness cannot always be guaranteed.

\begin{table}[t]
    \setlength{\tabcolsep}{5pt}
    \caption{\textbf{Robustness to perceptual degradation.} We reduce the number of input RGB-D views from 12 to 6, increasing the angular interval between observations to simulate degraded visual perception. Evaluation metrics reflect each model’s ability to navigate under reduced sensory input.}
    \label{tab:robust-perception}
    \centering
    \begin{tabular}{cccccccc}
        \toprule
        Method & Views & TL & NE$\downarrow$ & NDTW$\uparrow$ & OSR$\uparrow$ & SR$\uparrow$ & SPL$\uparrow$ \\
        \midrule
        \multirow{3}{*}{Open-Nav-GPT4} & 12 & 7.68 & 6.70 & 45.79 & 23 & 19 & 16.10 \\
         & 6 & 8.25 & 8.07 & 40.29 & 19 & 10 & 8.58 \\
         & & \textcolor{green}{$\uparrow$7.44\%} & \textcolor{green}{$\uparrow$20.45\%} & \textcolor{green}{$\downarrow$12.02\%} & \textcolor{green}{$\downarrow$17.39\%} & \textcolor{green}{$\downarrow$47.37\%} & \textcolor{green}{$\downarrow$46.73\%} \\
        \midrule
        \multirow{3}{*}{STRIDER} & 12 & 8.13 & 6.91 & 51.87 & 39 & 35 & 30.30 \\
         & 6 & 8.26 & 7.83 & 47.98 & 35 & 23 & 20.47 \\
         & & \textcolor{green}{$\uparrow$1.60\%} & \textcolor{green}{$\uparrow$13.33\%} & \textcolor{green}{$\downarrow$7.49\%} & \textcolor{green}{$\downarrow$10.26\%} & \textcolor{green}{$\downarrow$34.29\%} & \textcolor{green}{$\downarrow$32.47\%} \\
        \bottomrule
    \end{tabular}
\end{table}

\begin{table}[t]
    \caption{\textbf{Robustness to perturbation recovery}. Evaluation metrics are measured after injecting a mid-trajectory perturbation that displaces the agent from its planned waypoint, assessing its ability to recover and complete the navigation task.}
    \label{tab:robust-perturbation}
    \centering
    \begin{tabular}{ccccccc}
        \toprule
        Method  & TL & NE$\downarrow$ & NDTW$\uparrow$ & OSR$\uparrow$ & SR$\uparrow$ & SPL$\uparrow$ \\
        \midrule
        Open-Nav-GPT4  & 8.95 & 8.32 & 39.26 & 15 & 13 & 10.24 \\
        STRIDER        & 8.47 & 7.52 & 47.62 & 31 & 28 & 23.88 \\
        \bottomrule
    \end{tabular}
\end{table}

\subsection{Robustness to Perturbation Recovery}

To validate the robustness of STRIDER in recovering from unexpected execution deviations, we design a perturbation recovery experiment that simulates real-world disturbances during navigation. Specifically, at a randomly selected step during each episode, the agent's current waypoint is perturbed by altering its location with a slight displacement from the original path. This setting aims to evaluate whether the agent can detect the deviation, reorient itself, and still accomplish the navigation goal effectively.

As shown in Table~\ref{tab:robust-perturbation}, both {Open-Nav-GPT4} and {STRIDER} experience performance degradation due to mid-trajectory perturbations. However, STRIDER demonstrates significantly stronger robustness across all evaluation metrics. In terms of trajectory length (TL), Open-Nav-GPT4 exhibits a more pronounced increase (from 7.68 to 8.95) compared to STRIDER (from 8.13 to 8.47), indicating that STRIDER recovers more efficiently with fewer detours. Similarly, navigation error (NE) increases to 8.32 for Open-Nav-GPT4 but remains notably lower for STRIDER at 7.52, suggesting better spatial correction and path realignment capabilities.

When examining semantic alignment metrics, STRIDER retains a higher normalized dynamic time warping (NDTW) score of 47.62, compared to 39.26 for Open-Nav-GPT4, reflecting its superior ability to return to semantically plausible trajectories after deviation. Success-related metrics further highlight this gap: STRIDER achieves a 28\% success rate (SR) post-perturbation, more than double that of Open-Nav-GPT4 (13\%). The success weighted by path length (SPL) shows a similar trend, with STRIDER maintaining 23.88 versus Open-Nav-GPT4’s 10.24, confirming that not only does STRIDER recover better, but it also does so more efficiently.

Overall, these results underscore the importance of robust recovery mechanisms in navigation agents. While both models struggle under unexpected disruptions, STRIDER’s relatively graceful degradation suggests that it possesses stronger global scene grounding and re-planning capabilities. This robustness is critical for real-world deployment, where perfect plan execution cannot be assumed, and systems must dynamically adapt to noisy or unstable environments.

\begin{figure*}[t]
  \centering
  \includegraphics[width=1\linewidth]{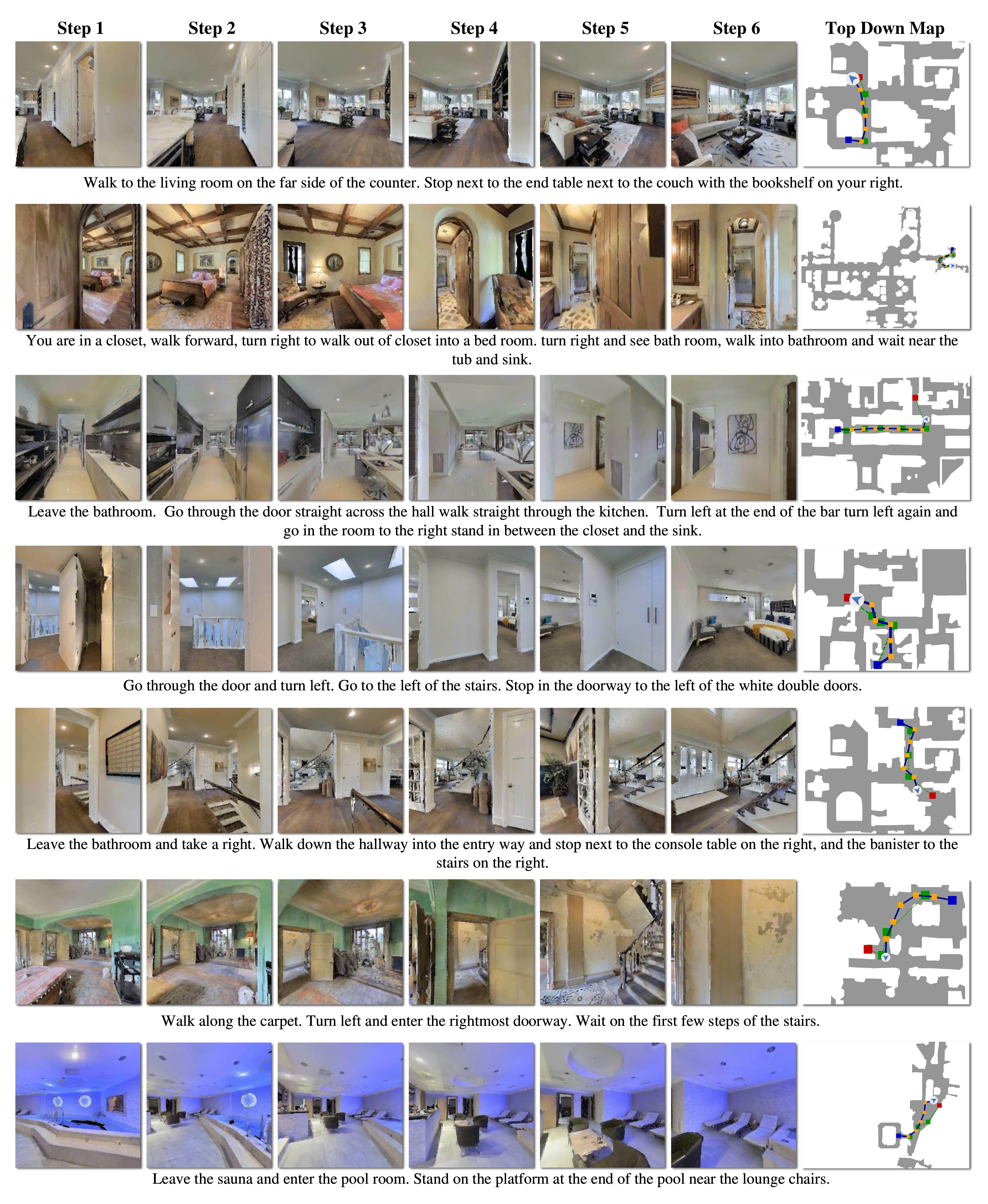}
  \caption{\textbf{Qualitative trajectory visualizations on the R2R-CE dataset.} Each row represents an instruction-following episode, where images from left to right correspond to the agent’s visual observations over time. The green line denotes the reference path, the blue dot indicates the agent's starting position, and the red dot marks the target location.}
  \label{fig:morer2r}
\end{figure*}

\section{Visualization Results}
\label{sec:visualization}

To complement the quantitative evaluation, we present a series of qualitative visualizations that offer deeper insight into the agent’s behavior, perception, and reasoning process. These include a detailed case study, additional trajectory examples on benchmark datasets, and representative failure cases. 

\subsection{Case Study}

\begin{figure*}[h]
  \centering
  \includegraphics[width=1\linewidth]{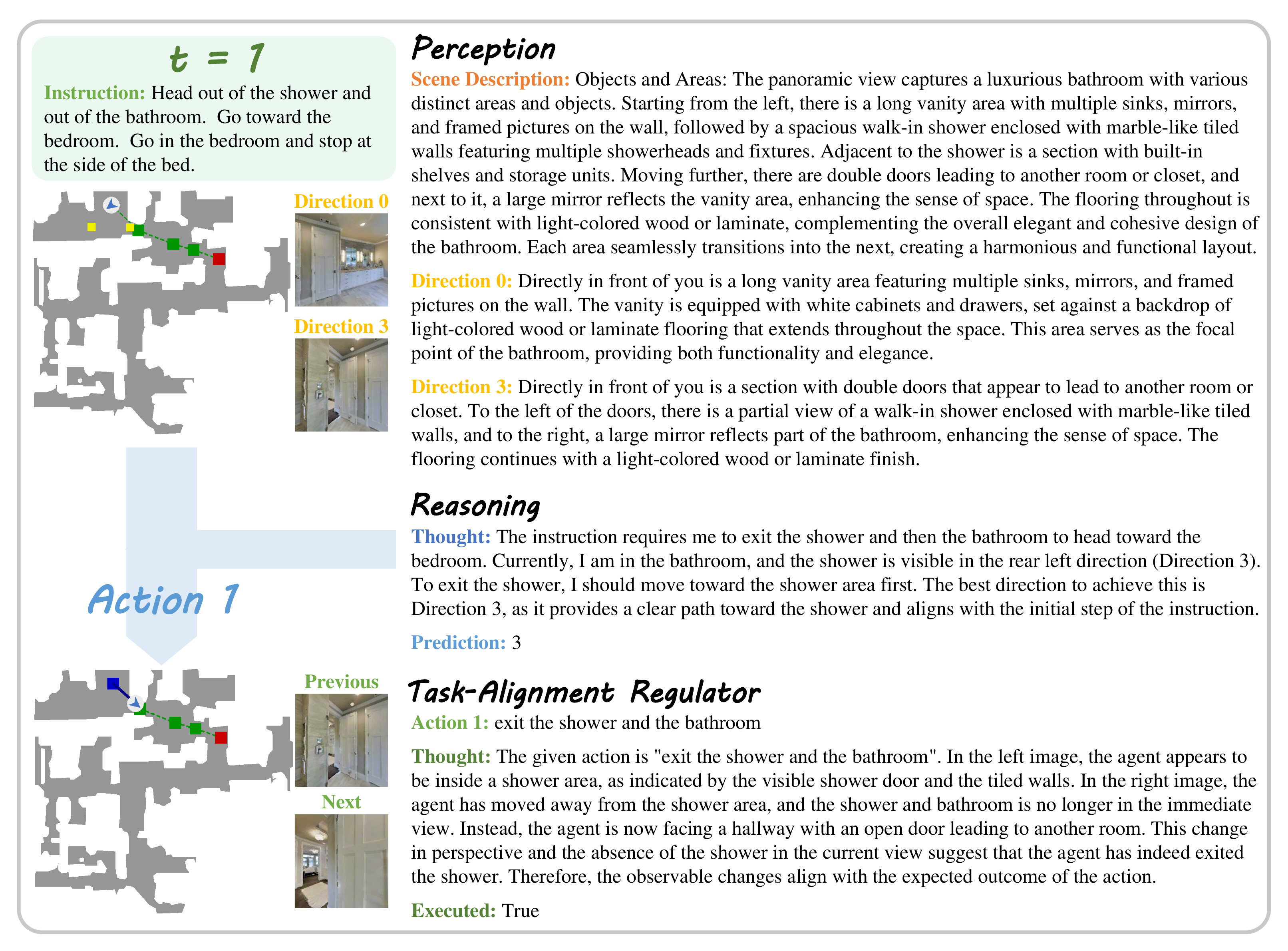}
  \caption{\textbf{Case study} with the navigation process at step 1.}
  \label{fig:case1}
\end{figure*}

\begin{figure*}[t]
  \centering
  \includegraphics[width=1\linewidth]{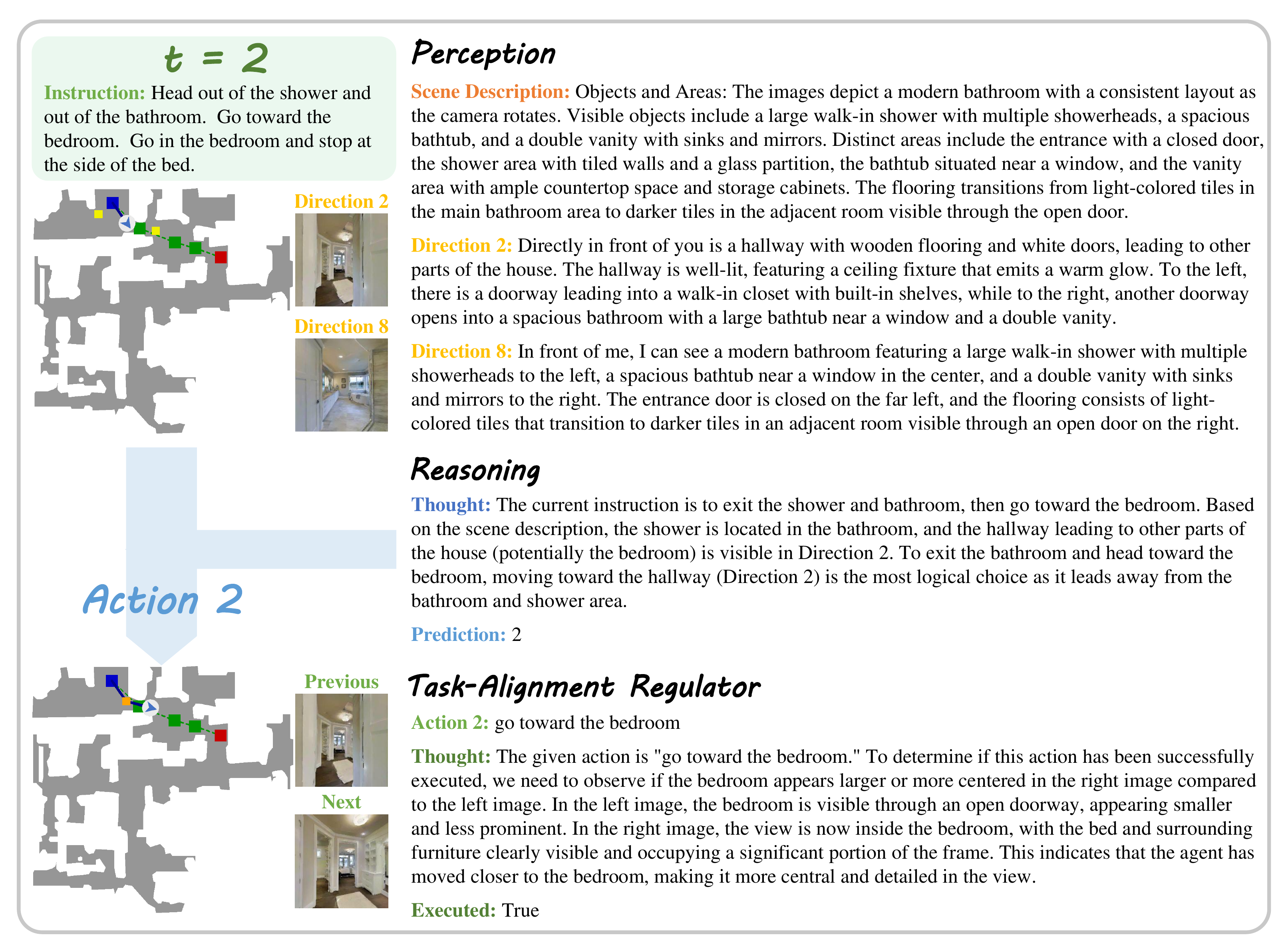}
  \caption{\textbf{Case study} with the navigation process at step 2.}
  \label{fig:case2}
\end{figure*}

\begin{figure*}[t]
  \centering
  \includegraphics[width=1\linewidth]{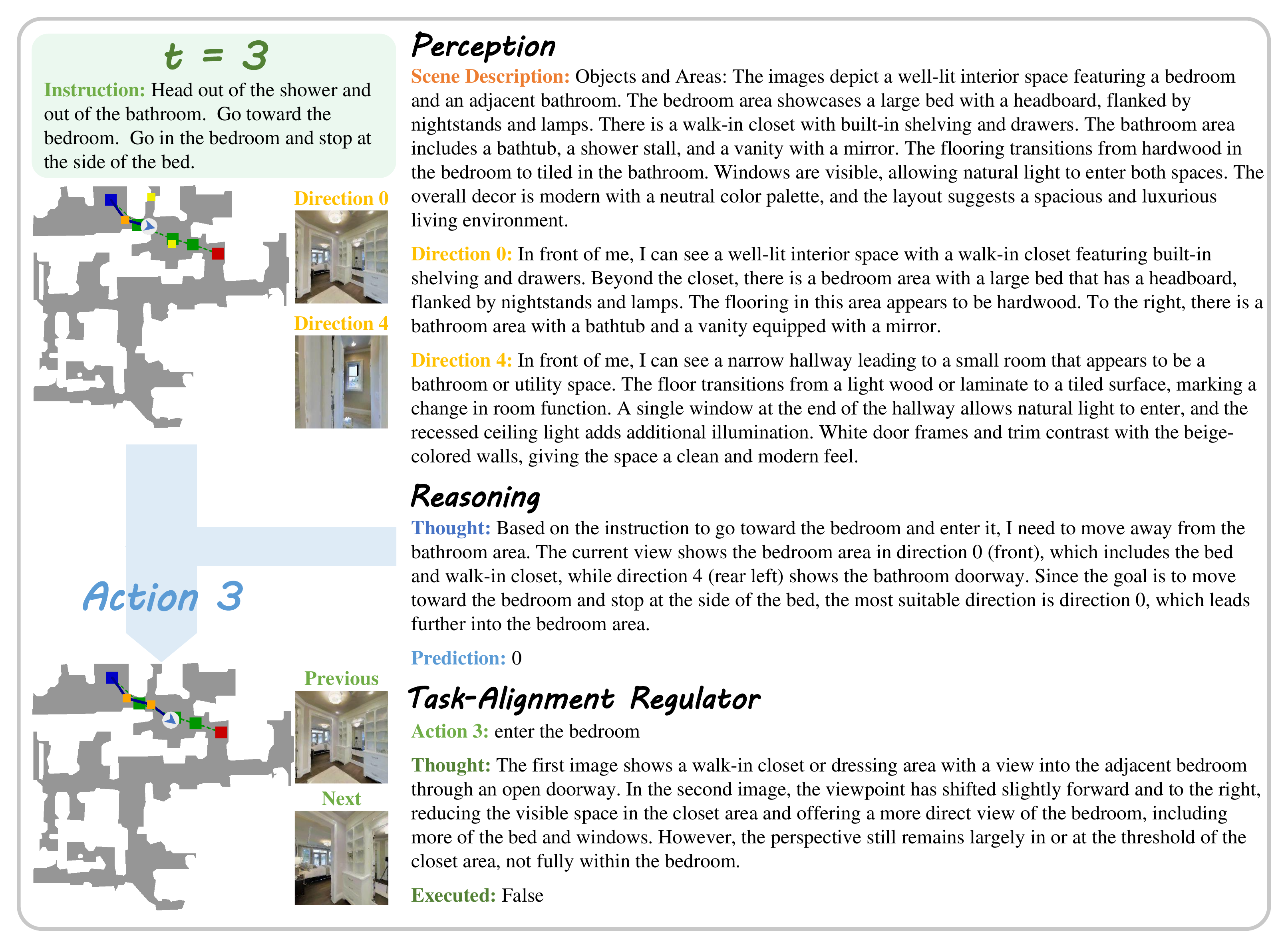}
  \caption{\textbf{Case study} with the navigation process at step 3.}
  \label{fig:case3}
\end{figure*}

\begin{figure*}[t]
  \centering
  \includegraphics[width=1\linewidth]{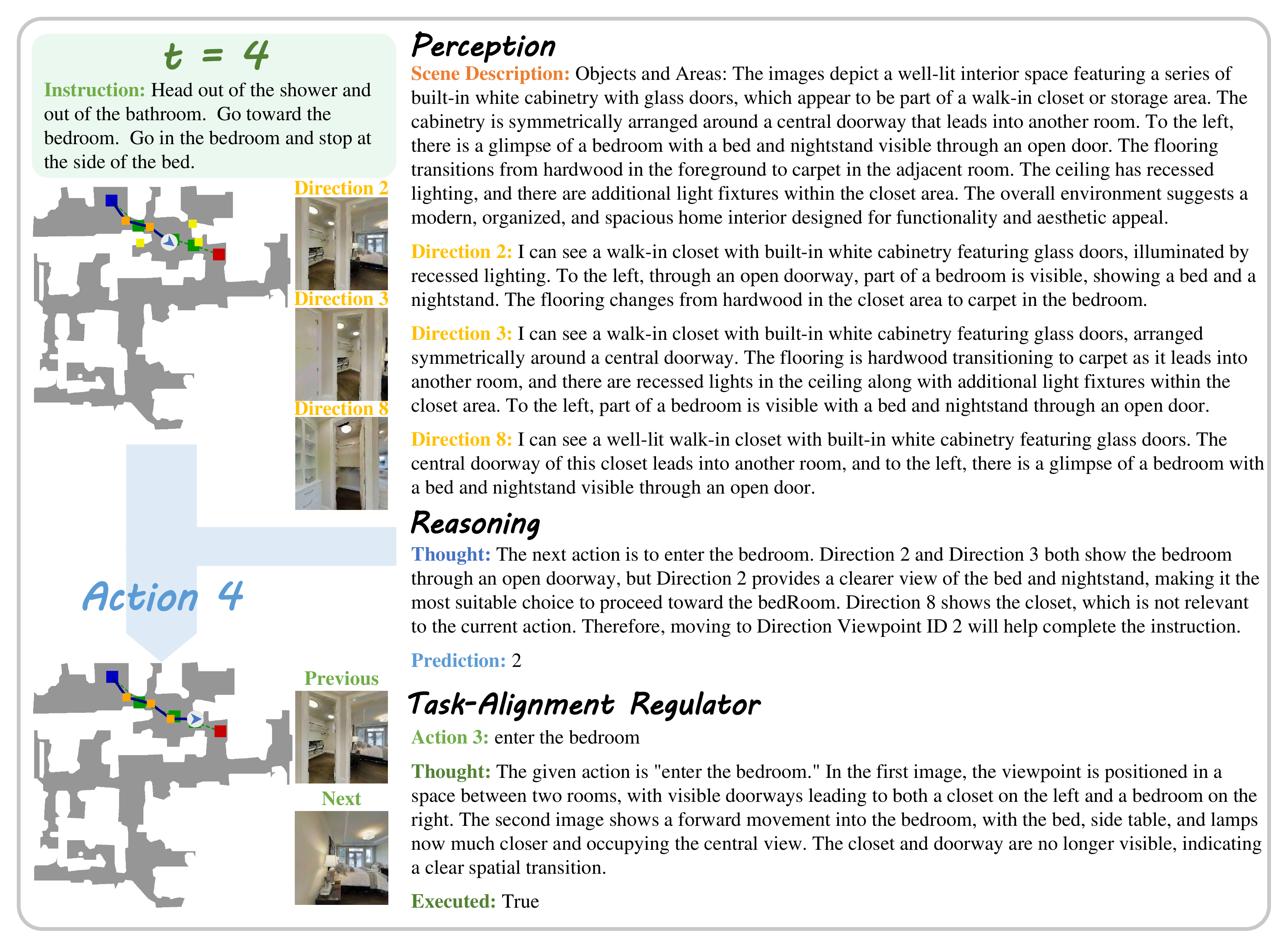}
  \caption{\textbf{Case study} with the navigation process at step 4.}
  \label{fig:case4}
\end{figure*}

\begin{figure*}[t]
  \centering
  \includegraphics[width=1\linewidth]{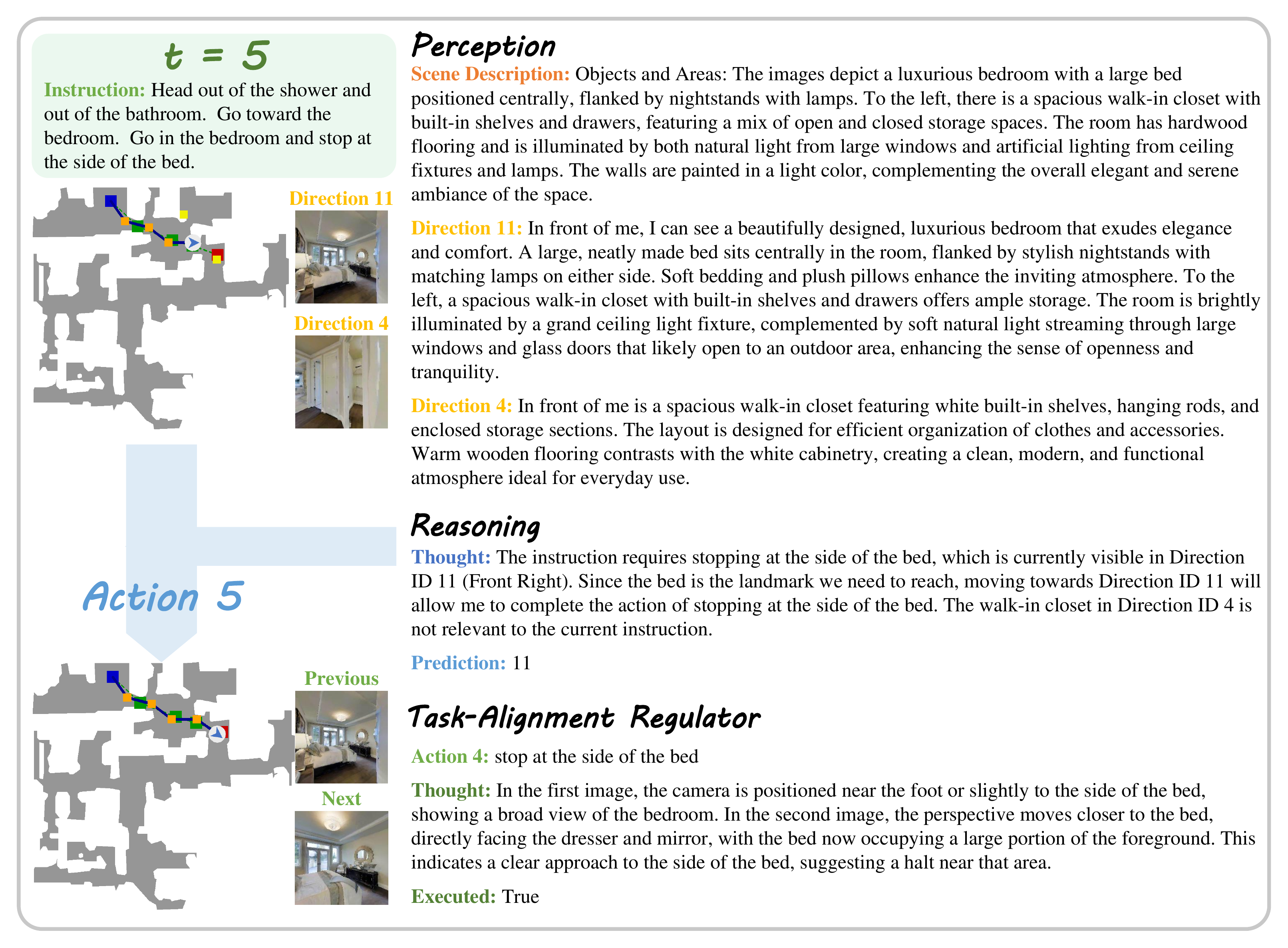}
  \caption{\textbf{Case study} with the navigation process at step 5.}
  \label{fig:case5}
\end{figure*}

To better understand how STRIDER perceives its environment, interprets instructions, and makes decisions throughout a navigation episode, we conduct a detailed case study with full-step visualization. As shown in Fig.~\ref{fig:case1}--\ref{fig:case5}, we select a representative trajectory and visualize each decision step, from $t=1$ to $t=5$. At each step, we include the panoramic scene description, directional candidate views, the agent's reasoning process, selected action, and the task-alignment feedback generated after execution.

This case illustrates how the agent sequentially decomposes the instruction, grounds it to visual observations, and dynamically reasons over available waypoints. The visualizations provide insights into the agent’s multimodal understanding, as well as its ability to monitor task progress and adjust behavior based on feedback. This step-by-step breakdown offers a qualitative complement to the quantitative results, highlighting the interpretability of our STRIDER.

\begin{figure*}[t]
  \centering
  \includegraphics[width=1\linewidth]{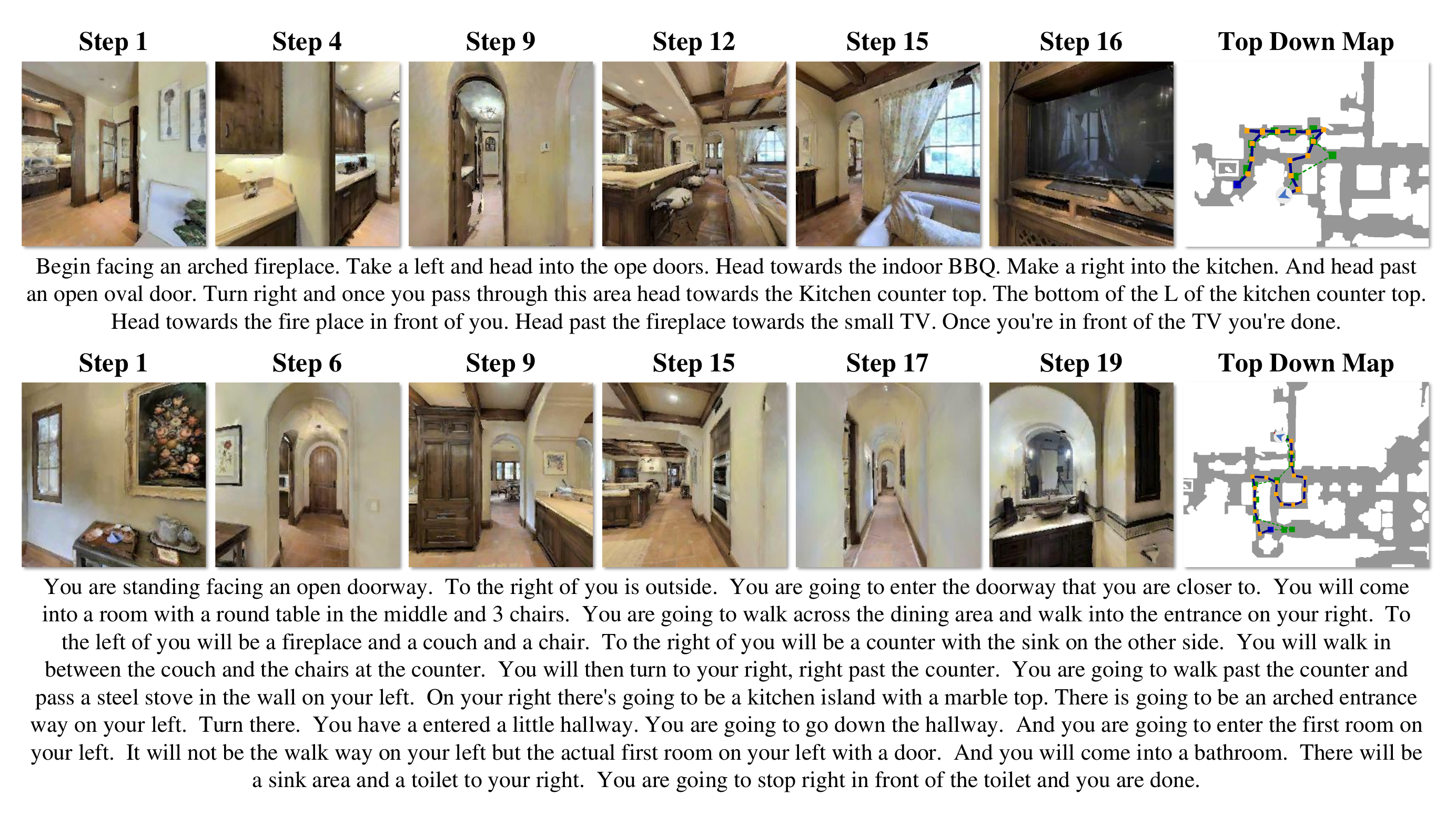}
  \caption{\textbf{Qualitative trajectory visualizations on the RxR-CE dataset.} Each row represents an instruction-following episode, where images from left to right correspond to the agent’s visual observations over time. The green line denotes the reference path, the blue dot indicates the agent's starting position, and the red dot marks the target location.}
  \label{fig:morerxr}
\end{figure*}

\subsection{More Visualization Trajectories}

To further illustrate the agent’s navigation behavior, we present additional qualitative trajectory visualizations on both the R2R-CE and RxR-CE datasets, as shown in Fig.~\ref{fig:morer2r} and Fig.~\ref{fig:morerxr}. For each selected episode, we visualize the agent’s step-by-step observations in temporal order by arranging the visual input images from left to right. This provides a clear view of how the agent perceives and interprets its environment over time. In addition, we include top-down map visualizations showing both the agent's executed trajectory (\textcolor{blue}{Blue Line}) and the annotated reference path (\textcolor{green}{Green Line}). The starting position and target location are marked with blue and red dots, respectively. These visualizations provide insights into the agent’s decision-making dynamics, spatial understanding, and alignment with human demonstration trajectories.

\subsection{Failure Cases}

\begin{figure*}[t]
  \centering
  \includegraphics[width=1\linewidth]{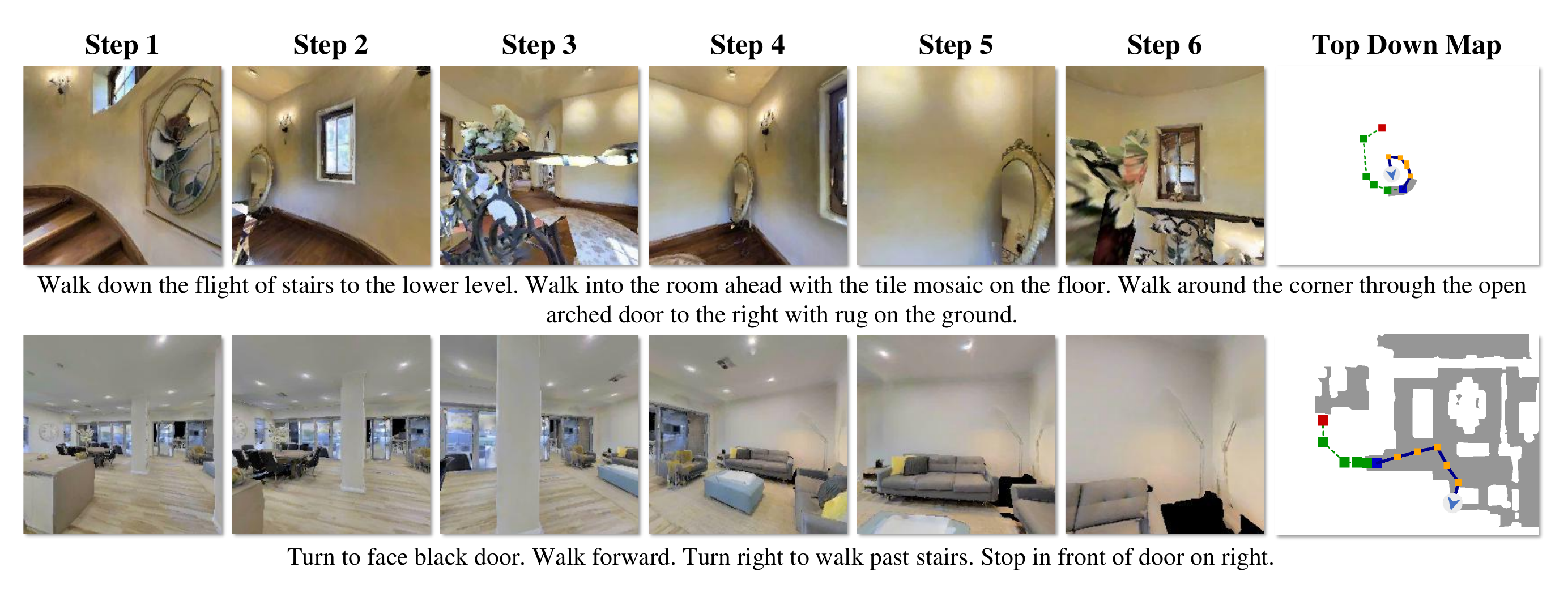}
  \caption{\textbf{Failure case caused by limited visual coverage and VLM misperception.} The agent only receives horizontally aligned images and is unable to perceive the staircase below. In addition, the VLM fails to detect the black door referenced in the instruction, leading to incorrect navigation.}
  \label{fig:failure}
\end{figure*}

Despite the overall effectiveness of STRIDER, we observe certain failure cases that reveal limitations imposed by the perception setting, as shown in Fig.~\ref{fig:failure}. The agent is instructed to descend a staircase and proceed toward a designated door. During navigation, the agent only receives horizontally aligned RGB-D inputs and cannot perceive content above or below its eye level according to the perception setting. As a result, the downward staircase is not captured in its visual input, leading to a misinterpretation of the scene and incorrect action selection.
Additionally, the VLM fails to recognize the “black door” referenced in the instruction during the early stages of execution. This semantic grounding error contributes further to the deviation from the intended trajectory. Such cases highlight the importance of vertical field-of-view coverage and more accurate visual grounding when operating in spatially complex environments.

\section{Limitations and Future Work}
\label{sec:limitation}

While STRIDER demonstrates promising performance, there remain several limitations associated with the system setting and underlying model capabilities. First, due to the perception setup, the agent only receives RGB-D observations captured at a fixed horizontal viewing angle. This configuration excludes upward or downward tilt, leading to incomplete vertical field-of-view coverage. In environments with significant elevation changes, such as staircases or split-level layouts, this limitation may result in missing important visual cues that are outside the visible range.
Second, our system relies heavily on the capabilities of large-scale VLMs and LLMs. While these models offer powerful generalization across domains, they also exhibit inherent weaknesses. In particular, VLMs may fail to consistently identify task-relevant objects or overlook subtle visual cues due to their goal-agnostic training paradigm. Similarly, LLMs may occasionally generate plausible yet incorrect decisions when provided with ambiguous or incomplete visual descriptions. Since our method builds on top of these pretrained models, its effectiveness is inherently constrained by their perceptual and reasoning limitations.
Lastly, the use of large-scale foundation models introduces additional computational latency. The combined inference time of VLM and LLM modules may limit the system’s applicability in real-time or high-frequency decision-making scenarios.
In future work, we aim to explore more perception-rich configurations that include vertical view coverage and investigate more lightweight, powerful VLM variants to improve both perceptual relevance and runtime efficiency.

\section{Broader Impacts}
\label{sec:impact}

Our work explores the integration of large-scale vision-language models (VLMs) and language models (LLMs) for embodied navigation, contributing to more generalizable, interpretable, and instruction-following agents in complex 3D environments. This approach has the potential to enhance human-robot interaction in practical scenarios such as indoor service robotics, assistive navigation, and autonomous exploration. By leveraging multimodal understanding and reasoning, our method brings AI agents closer to natural, language-driven task execution.
However, the reliance on large pre-trained models also introduces concerns regarding bias, reliability, and computational accessibility. VLMs may inadvertently encode cultural or dataset-specific biases, which could affect perception or decision-making in sensitive environments. Moreover, the high inference cost may limit deployment in resource-constrained or latency-sensitive settings. We encourage future research to investigate fairer, more efficient, and human-aligned multimodal systems, especially in safety-critical applications.


\clearpage

\end{document}